\documentclass[sigconf]{acmart}
\usepackage{bm}    
\usepackage{multirow}

\usepackage{amssymb}

\usepackage{amsmath}
\usepackage{colortbl}
\usepackage{xcolor} 
\usepackage{graphicx}
\usepackage{tcolorbox}
\usepackage{array}
\usepackage{booktabs}
\usepackage{multirow}
\usepackage{cleveref}
\usepackage{xspace}  
\usepackage{booktabs}

\begin{document}

\title{Bridging Large Language Models and Graph Structure Learning Models for Robust Representation Learning}
\newcommand{\mname}{LangGSL\xspace}
\newcommand{\gslm}{GSLM\xspace}






\author{Guangxin Su}
\affiliation{%
  \institution{The University of New South Wales}
  \city{Sydney}
  \country{Australia}}
\email{guangxin.su@unsw.edu.au}

\author{Yifan Zhu}
\affiliation{%
  \institution{The University of New South Wales}
  \city{Sydney}
  \country{Australia}}
\email{yifan.zhu4@student.unsw.edu.au}

\author{Wenjie Zhang$^{*}$}
\affiliation{%
  \institution{The University of New South Wales}
  \city{Sydney}
  \country{Australia}}
\email{wenjie.zhang@unsw.edu.au}
\thanks{$^*$Corresponding author.}

\author{Hanchen Wang}
\affiliation{%
  \institution{The University of New South Wales}
  \city{Sydney}
  \country{Australia}}
\email{hanchen.wang@unsw.edu.au}
\thanks{The code will be available after the review process. 
For early access, you may request it from the first author or the corresponding author with a reasonable justification, thank you for your understanding.}

\author{Ying Zhang}
\affiliation{%
  \institution{Zhejiang Gongshang
University}
  \city{Hangzhou}
  \country{China}}
\email{ying.zhang@zjgsu.edu.cn}

\renewcommand{\shortauthors}{Su et al.}

\begin{abstract}
Graph representation learning, involving both node features and graph structures, is crucial for real-world applications but often encounters pervasive noise. 
State-of-the-art methods typically address noise by focusing separately on node features with large language models (LLMs) and on graph structures with graph structure learning models (GSLMs). 
In this paper, we introduce \mname, a robust framework that integrates the complementary strengths of pre-trained language models and GSLMs to jointly enhance both node feature and graph structure learning.
In \mname, we first leverage LLMs to filter noise in the raw data and extract valuable cleaned information as features, enhancing the synergy of downstream models.
During the mutual learning phase in \mname, the core idea is to leverage the relatively small language model (LM) to process local attributes and generate reliable pseudo-labels and informative node embeddings, which are then integrated into the GSLM's prediction phase. 
This approach enriches the global context and enhances overall performance.
Meanwhile, GSLM refines the evolving graph structure
constructed from the LM's output, offering updated labels back to the LM as additional guidance, thus facilitating a more effective mutual learning process. 
The LM and GSLM work synergistically, complementing each other’s strengths and offsetting weaknesses within a variational information-maximizing framework, resulting in enhanced node features and a more robust graph structure.
Extensive experiments on diverse graph datasets of varying scales and across different task scenarios demonstrate the scalability and effectiveness of the proposed approach.
\end{abstract}



\keywords{Graph structure learning, Graph representation learning, Large language models, Graph neural networks}

\received{20 February 2007}
\received[revised]{12 March 2009}
\received[accepted]{5 June 2009}

\maketitle
\section{Introduction}
Graph Neural Networks (GNNs) \cite{kipf2016semi, velivckovic2017graph, hamilton2017inductive, xu2018powerful, yun2019graph} have emerged as the de facto method for modeling graph-structured data, which is increasingly prevalent and essential in numerous real-world applications.
Mainstream GNN methods implicitly propagate, aggregate, and preserve both node features and the graph structure, whose mutual influence is crucial for achieving optimal performance.
However, the noise and shallowness of node features, along with frequently suboptimal node connections, both stemming from the complexity of graph formation, present significant challenges to achieving accurate and robust graph modeling.
These challenges become particularly notable in various real-world applications. 
For instance, in social networks and citation networks \cite{quan2023robust, wu2023decor}, the graph structure is often noisy or incomplete due to the presence of spurious connections, missing links, or dynamic changes in graph structures over time.
In some cases, such as biological interaction networks \cite{yuan2024inferring, zhang2021synergistic, su2024inferring}, the graph structure may even be entirely unobserved, necessitating its inference from data.
Furthermore, the node embeddings derived from native data are often noisy \cite{zhu2024robust, luan2024graph, zhang2024integrating}, suffering from issues such as unstructured formatting, and variations across different platforms.

Recent efforts have increasingly focused on enhancing graph representation learning by leveraging large language models (LLMs), particularly when dealing with diverse modalities of data associated with graph nodes \cite{liu2023one, zhang2022protein, chen2024survey, theodoris2023transfer, schafer2024integrating}.
In this work, we primarily focus on graph representation learning on text-attributed graphs (TAGs) \cite{yang2021graphformers}, where nodes are associated with text entities.

\noindent \textbf{\textit{Motivation 1: }Combining deep text embeddings generated by LLMs with GNNs.}
Building on the foundations of LLMs, existing works on TAGs \cite{he2023harnessing, chen2024exploring, huang2024can, jin2023patton, zhao2024all} have demonstrated the effectiveness in node classification task.
On the one hand, LLMs ($e.g.$, GPT-4 \cite{achiam2023gpt}) possess advanced language processing capabilities, including complex reasoning and zero-shot learning.
Several works \cite{he2023harnessing, chen2024llaga, huang2024can} utilize the explanations of LLMs as distilled features for a downstream GNN based on task-relevant prompts, resulting in a significant performance improvement.
On the other hand, language models (LMs\footnote{The comparison between LMs and LLMs reveals a trade-off in capability and flexibility. LLMs, with their significantly larger parameter sizes, demonstrate enhanced capabilities and a vast repository of knowledge. However, this increased power often comes at the expense of flexibility. In contrast, LMs, though less capable, are more compact and thus can be fine-tuned more efficiently for specific tasks. \cite{he2023harnessing, chen2024exploring}})
can be effectively fine-tuned with a limited amount of labeled data to generate informative node embeddings for downstream tasks \cite{liu2023one, jin2023patton, chien2021node}.
The common idea behind these works is to leverage the strengths of pre-trained language models to enhance node representations and achieve improved performance; however, they still face inherent limitations.
\underline{\textit{Limitation 1: }}Existing LLM-based methods primarily rely on well-structured raw data ($e.g.$, titles and abstracts serve as node attributes in citation networks). However, it remains unclear whether LLMs have the capability to effectively clean noisy node attributes.
\underline{\textit{Limitation 2: }}LLM-based graph representation learning methods typically depend on a fixed graph structure that remains static during training,  making them incapable of updating noisy structures and handling scenarios where the graph structure is absent.

\noindent \textbf{\textit{Motivation 2: } Drawing insights from existing graph structure learning methods.}
The exploration of learning robust graph structures using vanilla GNNs has yielded extensive insights.
Specifically, graph structure learning models (GSLMs) are crucial in addressing the challenges of modeling real-world graphs, which are often redundant, biased, noisy, incomplete, or even unavailable.
An ideal approach should flexibly integrate these methods within a modular framework, enabling adjustments to varying scenarios and graph characteristics.
However, several critical limitations still need to be addressed to fully realize this potential.
\underline{\textit{Limitation 3: }}{The model may encounter significant scalability issues when applied to large graphs.}
GSLMs~\cite{franceschi2019learning, jin2020graph, guo2024graphedit, liu2022compact, wang2021graph} necessitate evaluating all potential node pair relationships, resulting in a proportional increase in both computational demands and memory requirements for the graph structure optimization, particularly in large-scale graphs.
Additionally, these memory costs will be unaffordable in real-world TAGs where nodes are densely connected.
\underline{\textit{Limitation 4:} }{
The initial node features in existing GSLMs \cite{chen2020iterative, li2022reliable, wu2022nodeformer, wu2023homophily} are often shallow and noisy, usually derived from noisy native data using simplistic embedding techniques like Bag-of-Words \cite{harris1954distributional} for text data, which limits performance from the outset.}

\begin{figure}[t!]
    \centering
    \includegraphics[width=\linewidth]{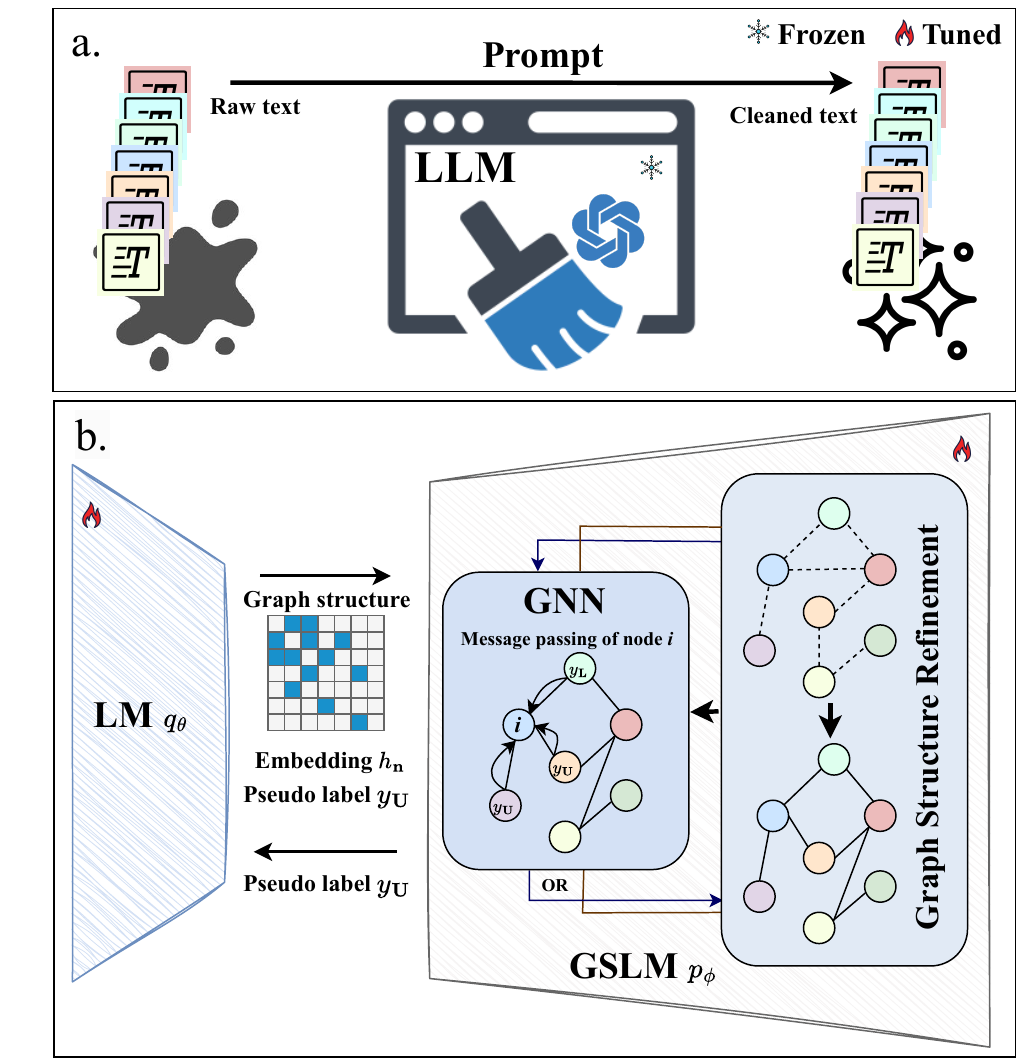}
    \caption{The pipeline of \mname: (a), LLM filters out irrelevant or noisy information from raw text and provides task-related text attributes using designed prompts.
    (b), An iterative optimization process occurs between the LM and the graph structure learning model (GSLM), where the LM generates graph structures, embeddings, and pseudo labels based on cleaned text attributes provided by the LLM. In turn, the GSLM refines the graph structure and provides updated pseudo labels back to the LM. 
    Three interaction mechanisms are introduced between the vanilla GNN and graph structure refinement component in GSLM.}
    \label{fig:framework}
\end{figure}

\noindent \textbf{\textit{Motivation 3: }Enabling mutual learning between language models and GSLMs. 
}To leverage the strengths and address the limitations of pre-trained language models and GSLMs, we propose a \textbf{Lang}uage model-enhanced \textbf{G}raph \textbf{S}tructure \textbf{L}earning framework, named \textbf{\mname}.
In \mname, the LLM ($e.g.$, GPT-3.5-turbo \cite{ouyang2022training}) is first applied to extract relevant prior knowledge and organize the noisy raw texts into a more structured format using task-related prompts (Figure \ref{fig:framework}a),  making the generated text attributes digestible and informative for training smaller models.
During the mutual training phase, we iteratively train both the LM to enrich local textual features and the \gslm part to refine global graph structure information (Figure \ref{fig:framework}b).
The LM and GSLM primarily interact by exchanging pseudo labels, facilitating the alignment of local and global label distributions, and are jointly optimized within an evidence lower bound framework for improved performance.
In addition, the LM takes the cleaned text attributes generated by LLM as input and updates the node embeddings through a fine-tuning process. 
Both the enhanced node embeddings and the initial graph structure, derived from these embeddings, are then transferred to the \gslm module for further processing.
The \gslm module consists of two flexible components: a vanilla GNN and a graph structure refinement module, both of which can be adapted and utilized as needed.
The graph structure refinement module aims to identify an implicit graph structure that enhances the initial one. 
This refined structure is then passed to the vanilla GNN, allowing both components in GSLM to be jointly optimized for graph structure learning.
Notably, as evidenced by experiments, based on the cleaned and enhanced node embeddings, the \gslm module using only a vanilla GNN demonstrates performance comparable to baseline methods while being more efficient.

\begin{figure}[t!]
    \centering
    \includegraphics[width=0.8\linewidth]{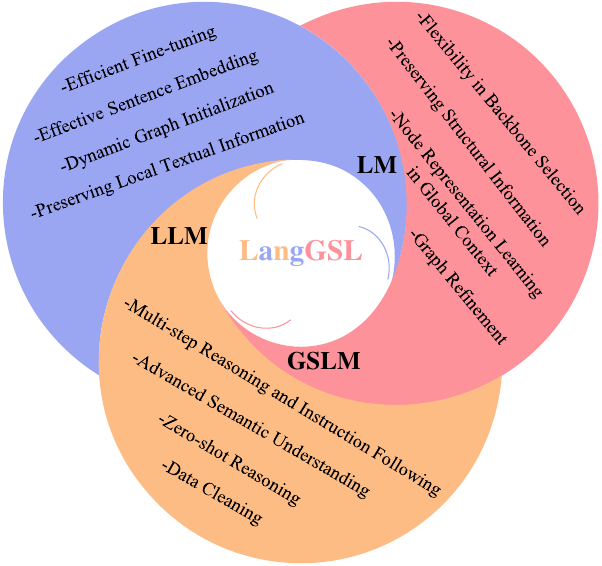}
    \caption{Key advantages of modules in the \mname. }
    \label{fig:advan}
\end{figure}

\noindent \textbf{\textit{Contributions:}}
\begin{itemize}
\item We present LangGSL, a comprehensive framework that seamlessly integrates node feature learning and graph structure refinement, advancing the state-of-the-art in graph representation learning.
\item We develop a flexible mechanism that allows language models and GSLMs to mutually enhance performance by aggregating their respective advantages (Figure \ref{fig:advan}), through minimizing a variational task-specific loss and incorporating a graph refinement loss.
\item We conduct comprehensive experimental evaluations, and the results demonstrate the superior performance of our approach across diverse task scenarios, including noisy or attacked graph structures, cases where the graph structure is absent, and other challenging conditions.

\end{itemize}



\section{Related works}
\subsection{Graph Structure Learning}
The problem of graph structure learning has been widely studied in the context of GNN from different perspectives.
The majority of the graph structure learning methods still adopt a co-training approach \cite{franceschi2019learning, chen2020iterative, wu2022nodeformer, yu2021graph, jin2020graph, fatemi2021slaps, zhang2024curriculum}, where the graph structure is optimized jointly with the entire neural network.
For instance, IDGL \cite{chen2020iterative} jointly and iteratively learns the graph structure and graph embeddings, using the associated learning metric as the optimization objective.
RCL \cite{zhang2024curriculum} first extracts latent node embeddings using a GNN model based on the current training structure, then jointly learns the node prediction labels and reconstructs the input structure through a decoder.
An alternative approach to graph structure learning involves using a dedicated module to refine the graph structure, which is subsequently passed to downstream GNNs for further processing \cite{wang2021graph, liu2022compact, zou2023se}.
Within the category, SEGSL \cite{zou2023se} introduces graph structural entropy, optimizing the structure using an one-dimensional entropy maximization strategy, constructing an encoding tree to capture hierarchical information, and subsequently reconstructing the graph from this tree, which is then incorporated into a GNN.
Iterative methods \cite{wang2021graph, song2022towards} train the two components in cycles, learning the graph structure from predictions or representations produced by an optimized GNN, which is then used to train a new GNN model in the next iteration.
While existing graph structure learning methods have shown promise, they rely heavily on explicit graph structural information as supervisory signals, leaving them susceptible to issues like data noise and sparsity.

\subsection{Large Language Models for Graphs}
Leveraging the extensive knowledge of large language models (LLMs) for graph data enhancement has been a focus of recent studies, with several exploring the integration of LLMs and Graph Neural Networks (GNNs).
One category (LLMs as enhancers) utilizes LLMs for node feature augmentation \cite{he2023harnessing, liu2023one, huang2023can, guo2023gpt4graph} while relying on the standard GNN mechanisms to update node features, incorporating the underlying graph topological information for better performance.
As an example, GIANT \cite{chien2021node} integrates graph topology information into language models (LMs) through the XR-Transformer \cite{10.5555/3540261.3540817}, and the enhanced node embeddings can be further used in various downstream tasks.
Alternatively, efforts to represent graphs linguistically and treat LLMs as predictors \cite{huang2024can, chen2024llaga, zhao2023graphtext, ye2023natural}  risk losing structural information, as these methods may struggle to effectively translate graph structures into natural language.
To harness the complementary strengths of LLMs and GNNs, GNN-LLM alignment approaches \cite{zhao2022learning, jin2023patton} utilize cross-training between the two models, with alignment strategies designed to ensure both effectively capture graph topology and enhance node embeddings.
In contrast to \mname, methods like GLEM \cite{zhao2022learning}, a representative alignment approach, focus solely on optimizing the model with noisy node embeddings, neglecting explicit graph structure learning, thereby limiting the potential performance and scope of downstream tasks.
Considering graph structure learning, GraphEdit \cite{guo2024graphedit} leverages an large language model (LLM) to model node relationships based on paired node text attributes. 
However, compared to LLM-based methods focused on single-node feature augmentation, GraphEdit exhibits increased computational complexity as the number of nodes grows, with only marginal gains in performance.
LLM4RGNN \cite{zhang2024can} employs LLMs to defend against adversarial attacks, but this strategy incurs a trade-off, leading to a slight reduction in overall accuracy.
In contrast, \mname delivers superior performance across kinds of downstream tasks, while simultaneously maintaining a robust graph structure.

\section{Method}
In this section, we outline the pipeline of \mname, as illustrated in Figure \ref{fig:framework}, specifically tailored for the node classification task, detailing each stage from data cleaning to models' mutual learning.


\subsection{Preliminaries}
\noindent
\textbf{Task definition. }
Formally, a text-attributed graph (TAG) can be represented as \(\mathcal{G} = (\mathcal{V}, \bm{A}, \{\bm{s}_n\}_{n \in \mathcal{V}})\), where \(\mathcal{V}\) is a set of \(n\) nodes, nodes are initially connected via adjacent matrix \(\bm{A} \in \mathbb{R}^{n \times n}\), and \(\bm{s}_n \in \mathcal{D}^{L_n}\) is a sequential text associated with node \(n \in \mathcal{V}\), with \(\mathcal{D}\) as the words or tokens dictionary, and \(L_n\) as the sequence length. 
Given a noisy graph input \(\mathcal{G}\), or in the case that only node tokens \(\bm{s}_n \in \mathcal{D}^{L_n}\) are provided, where the initialization of \(\bm{A}\) will depend on the chosen strategy ($e.g.,$ kNN \cite{fix1985discriminatory}), the deep graph learning problem addressed in this paper is to produce an optimized graph:
\begin{equation}
\mathcal{G}^{*}: (\bm{A}^{*}, \bm{X}^{*}) = \psi(\mathcal{G}),
\end{equation}
where the learned graph \(\mathcal{G}^{*}\) may contain learnable parameters ($i.e.$, tunable edges or node features) within its adjacency matrix $\bm{A^{*}}$ or feature matrix $\bm{X^{*}}$. 
\(\psi(\cdot)\) denotes the designed learning framework that optimizes both initial node features and the underlying graph structures. 
The candidate space for refined \(\bm{A}^{*}\) is denoted as \(\mathbb{A}\), while the candidate space for the optimized node feature matrix \(\bm{X}^{*} \in \mathbb{R}^{n \times F}\) is represented as \(\mathbb{X}\),
with respect to some {(semi-)supervised} downstream tasks on TAGs. 

\noindent
\textbf{Evidence Lower Bound. }
Based on the obtained graph $\mathcal{G}:= (\bm{A}, \{\bm{s}_n\}_{n \in \mathcal{V}})$, we aim to search for the optimal $\bm{A}^{*}$ and $\bm{X}^{*}$ within their respective candidate spaces $\mathbb{A}$ and $\mathbb{X}$ that maximize the likelihood of correctly predicting the observed labels $\bm{y}_L$ using proposed framework $\psi(\cdot)$ named \mname.
The process can be expressed as:

\begin{equation}
\max_{{\bm{A}^{*}} \in \mathbb{A}, {\bm{X}^{*}} \in \mathbb{X}} P_{\psi}(\bm{y}_L \mid \{\bm{s}_n\}_{n \in \mathcal{V}}, \bm{A}).
\end{equation} 

Notably, in node classification tasks, the label of a node is not independent but influenced by the labels of its neighboring nodes.
This means that the observed labels $\bm{y}_L$ are statistically dependent on the unobserved labels $\bm{y}_U$.
To utilize this dependence relationship, we introduce a variational distribution \(q(\bm{y}_U \mid \bm{s}_U)\) to approximate the true posterior distribution \(p(\bm{y}_U \mid \{\bm{s}_n\}_{n \in \mathcal{V}}, \bm{A}, \bm{y}_L)\).
Following this idea, we derive the Evidence Lower Bound (ELBO) as:
\begin{equation}
\scriptsize
\begin{split}
\label{eq:ELBO}
    \log p(\bm{y}_L \mid \{\bm{s}_n\}_{n \in \mathcal{V}}, \bm{A}) 
    &{\geq} \mathbb{E}_{q(\bm{y}_U \mid \bm{s}_U)} \left[ \log \frac{p(\bm{y}_L, \bm{y}_U \mid \{\bm{s}_n\}_{n \in \mathcal{V}}, \bm{A})}{q(\bm{y}_U \mid \bm{s}_U)} \right] \\
    &{=} \mathbb{E}_{q(\bm{y}_U \mid \bm{s}_U)} \left[ \log p(\bm{y}_L, \bm{y}_U \mid \{\bm{s}_n\}_{n \in \mathcal{V}}, \bm{A}) \right]\\
    & \quad {-} \mathbb{E}_{q(\bm{y}_U \mid \bm{s}_U)} \left[ \log q(\bm{y}_U \mid \bm{s}_U) \right].
\end{split}
\end{equation}

This ELBO provides a tractable lower bound for the original log-likelihood, enabling efficient optimization. By maximizing the ELBO with respect to both the variational distribution and the model parameters, we indirectly maximize the true log-likelihood \(\log p(\bm{y}_L \mid \{\bm{s}_n\}_{n \in \mathcal{V}}, \bm{A})\).
{The complete derivation of the ELBO is included in the Appendix \ref{app:ELBO} for further reference.}

\subsection{Data Cleaning based on LLMs}

Given the substantial cost of obtaining human-annotated texts, most approaches \cite{zhao2022learning, jin2023patton, chien2021node} rely on raw text data scraped from the web in TAGs learning, which is frequently contaminated with noise, including unstructured formatting, multilingual content,  task-irrelevant information, and non-standardized formats, all of which can hinder downstream tasks and model performance.
{Despite employing simple prediction and explanation-based filters \cite{huang2024can, he2023harnessing, liu2023one}, noise remains prevalent in the texts.}

To effectively mitigate the impact of noise in raw text data and extract concise, task-specific inputs for downstream LM, we design prompts that balance generality and specificity: broad enough to capture essential task-relevant indicators while being sufficiently targeted to filter out irrelevant or noisy information (Figure \ref{fig:framework}a). 
Unlike prior work, our method first instructs the LLM to summarize input text, including symbols like emojis, before performing classification and explanation tasks. This preprocessing step enables the LLM to handle discrete and symbolic language elements more effectively. Furthermore, we observed that LLMs tend to struggle with borderline classification cases, where the raw text is ambiguous. 
To address this issue, we provide key factors as additional instructions within the prompt, helping to guide the LLM towards more accurate and reliable predictions.
Our approach interacts exclusively via API access to an LLM, using text-based inputs and outputs, without requiring fine-tuning or direct access to model embeddings or logits.
These outputs from the LLM serve as supplementary text attributes for the downstream LMs and GSLMs.

Additional prompt samples, output comparisons between different LLMs, experimental validation of data cleaning effectiveness, and further discussions can be found in Appendix \ref{app:prompt}.

\subsection{Mutual Learning between LM and GSLM}
To leverage the complementary strengths of LM and GSLM, we jointly optimize the ELBO (\Cref{eq:ELBO}) and graph regularization through their interaction, enhancing both feature representation and structural learning (Figure \ref{fig:framework}b).
Specifically, the mutual learning involves two phases: (1) initializing the graph structure and optimizing  variational distribution $q$ by LM, and
(2) refining the graph structure within GSLM and optimizing the posterior distribution $p$ by GSLM, with GNN as the default backbone.
\subsubsection{\textbf{LM Learning Phase}}
\label{sec:MethodLM}
Leveraging the complex linguistic capabilities of LLMs and task-specific prompts, we obtain cleaned and task-relevant textual output $\bm{s}_n$ for node $n \in \mathcal{V}$. 
These text entities are then processed by a pre-trained $\text{LM}$, denoted as $q_{\theta}$ ($e.g.$, Sent-BERT \cite{reimers2019sentence}), which serves as an expert to refine the text attributes $\bm{s}_n$, generating and providing informative embeddings for \gslm. The resulting text embeddings are obtained as follows:

\begin{equation}
\bm{h}_n  =q_{\theta}{(\bm{s}_n)}\in \mathbb{R}^{d}.    
\end{equation}
\textbf{Initial graph structure formation. }Based on the inferred embeddings, the likelihood of a link between any two nodes can be quantified using a similarity measure function $Sim(\cdot)$ ($e.g.$, cosine similarity) between their respective embeddings $\bm{h}_i$ and $\bm{h}_v$, which
is as follows:
\begin{equation}
\label{eq:Sim()}
    f_{adj}(\bm{h}_i, \bm{h}_v)=Sim(\bm{h}_i, \bm{h}_v).
\end{equation}
Specifically, in each iteration, the graph structure is updated dynamically based on $Sim(\cdot)$ as the node embeddings are refined.
Alternatively, a cross-attention layer before the classification head can be employed to explicitly model the interactions between nodes,  where attention weights serve as an adaptive mechanism to capture the strength and relevance of relationships in a learnable way.

\noindent
\textbf{Optimization of distribution $q$.} Given the informative embeddings \( \bm{h}_n \), and the learned initial adjacency matrix \( \bm{A} \) based on the function ${Sim}(\cdot)$  (in cases where the initial graph structure \( \bm{A} \) is not available), the true posterior distribution can be represented as \( p_\phi(\bm{y}_U \mid \{\bm{h}_n\}_{n \in \mathcal{V}}, \bm{A}, \bm{y}_L) \).

In the LM learning phase of \mname, we will update the parameters of the LM \( q_\theta \), while keeping the \gslm parameters \( \phi \) fixed. 
Based on \Cref{eq:ELBO}, the goal of \mname is to refine the variational distribution \( q_\theta(\bm{y}_U \mid \bm{s}_U) \) so that it better approximates the true posterior distribution \( p_\phi(\bm{y}_U \mid \{\bm{h}_n\}_{n \in \mathcal{V}}, \bm{A}, \bm{y}_L) \) via minimizing the Kullback-Leibler (KL) divergence between \( q_\theta \) and $p_{\phi}$. 
By doing so, we enable the LM to capture global semantic correlations distilled from the \gslm.

However, directly minimizing the KL divergence between \( q_\theta \) and the true posterior is intractable due to the complexity of computing the entropy of the variational distribution.
Inspired by the ``wake-sleep'' algorithm \cite{hinton1995wake}, we instead opt to minimize the reverse KL divergence, which simplifies the optimization process.
Specifically, the optimization problem in the LM learning phase becomes:
\begin{align}
\min_\theta \ \mathrm{KL}\left( p_\phi(\bm{y}_U \mid \{\bm{h}_n\}_{n \in \mathcal{V}}, \bm{A}, \bm{y}_L), q_\theta(\bm{y}_U \mid \bm{h}_U)\right) \notag \\
= \max_\theta \ \mathbb{E}_{p_\phi(\bm{y}_U \mid \{\bm{h}_n\}_{n \in \mathcal{V}}, \bm{A}, \bm{y}_L)} \left[ \log q_\theta(\bm{y}_U \mid \bm{h}_U) \right].
\end{align}
as minimizing the KL divergence is equivalent to maximizing the expected log-likelihood of the LM under the posterior distribution estimated by the \gslm.
{In the LM, we assume that the labels of different nodes are determined solely by their respective text attributes, using a mean-field approximation \cite{gabrie2020mean} to capture this independence.}
This assumption leads to the following factorization, where the expectation is expressed as a sum over the unlabeled nodes:
\begin{equation} 
\max_\theta \ \sum_{m \in U} \mathbb{E}_{p_\phi(\bm{y}_m \mid \{\bm{h}_n\}_{n \in \mathcal{V}}, \bm{A}, \bm{y}_L)} \left[ \log q_\theta(\bm{y}_m \mid \bm{h}_m) \right]. 
\end{equation}

Here, directly computing the posterior distribution \( p_\phi(\bm{y}_m \mid \{\bm{h}_n\}_{n \in \mathcal{V}}, \bm{A}, \bm{y}_L) \),  conditioning solely on observed node labels $\bm{y}_L$ is insufficient, as unobserved labels from neighboring nodes are critical for capturing the full graph structure in GSLM. 
To approximate this distribution, we utilize the current LM to generate pseudo-labels for the unlabeled nodes \( U\setminus\{m\} \), and we can approximate the posterior distribution as:
\begin{equation} 
p_\phi\left(\bm{y}_m \mid \{\bm{h}_n\}_{n \in \mathcal{V}}, \bm{A}, \bm{y}_L, \hat{\bm{y}}_{U \setminus \{ m \}}\right), 
\end{equation}
where \( \hat{\bm{y}}_{U \setminus \{ m \}} = \{ \hat{\bm{y}}_{m'} \}_{m' \in U \setminus \{m\}} \) and each \( \hat{\bm{y}}_{m'} \) is sampled from \( q_\theta(\bm{y}_{m'} \mid \bm{h}_{m'}) \).

Including the labeled nodes in the training process, the overall objective function for updating the LM becomes:
\begin{align}
\label{eq:ELBO_LM}
\mathcal{L}_{\text{LM}} &= \alpha \sum_{m \in U} \mathbb{E}_{p_\phi\left(\bm{y}_m \mid \{\bm{h}_n\}_{n \in \mathcal{V}}, \bm{A}, \bm{y}_L, \hat{\bm{y}}_{U \setminus \{ m \}}\right)} \left[ \log q_\theta(\bm{y}_m \mid \bm{h}_m) \right] \notag \\
&\quad + (1 - \alpha) \sum_{m \in L} \log q_\theta(\bm{y}_m \mid \bm{h}_m),
\end{align}
where \( \alpha \in [0, 1] \) is a hyperparameter that balances the two terms. The first term represents knowledge distillation from the \gslm to the LM, encouraging the LM to align its predictions with the global structural information captured by the \gslm. 
The second term is a supervised learning objective based on the labeled nodes.
By optimizing this objective, the LM learns to produce label predictions that are consistent with both the textual content of each node and the structural patterns. 
Moreover, the graph structure inferred using \Cref{eq:Sim()} exhibits greater robustness as evidenced by experiments.

\begin{table*}[h]
\centering
\caption{Accuracy ± STD comparison (\%) under the standard setting of transductive node classification task in the Topology Refinement (TR) scenario, which means the original graph structure is
available for each method. 
The highest results are highlighted with \textbf{bold}, while the second highest results are marked with \underline{underline}. "OOM" denotes out of memory. 
"-" denotes that related prompts are not provided by the method.
}
\label{tab:gslWStr}
\begin{tabular}{lcccc}
\toprule
 & \multicolumn{1}{c}{Cora} & \multicolumn{1}{c}{Instagram} & \multicolumn{1}{c}{Pubmed} & \multicolumn{1}{c}{ogbn-arxiv} \\
Edge Hom. & 0.81 & 0.10 & 0.80 & 0.65 \\
\midrule
GCN & $87.36 \pm 1.60$ & $63.79 \pm 0.08$ & $78.98 \pm 0.35$ & $71.82 \pm 0.27$ \\
ProGNN & $85.79 \pm 0.37$ & $61.55 \pm 0.09$ & OOM & OOM \\
IDGL & $88.63 \pm 0.44$ & $64.67 \pm 0.32$ & $82.78 \pm 0.44$ & OOM \\
GEN & $86.53 \pm 0.63$ & $64.55 \pm 0.44$ & $78.91 \pm 0.69$ & OOM \\
LDS & $87.71 \pm 0.13$ & OOM & OOM & OOM \\
GRCN & $84.13 \pm 0.37$ & $63.81 \pm 0.37$ & $78.92 \pm 0.39$ & OOM \\
CoGSL & $82.07 \pm 0.51$ & $63.93 \pm 0.36$ & OOM & OOM \\
SUBLIME & $85.04 \pm 0.37$ & $60.33 \pm 0.90$ & $80.90 \pm 0.94$ & $71.75 \pm 0.36$ \\
STABLE & $88.75 \pm 0.35$ & $63.56 \pm 0.03$ & $81.46 \pm 0.78$ & OOM \\
NodeFormer & $88.56 \pm 1.01$ & $64.73 \pm 1.00$ & $81.51 \pm 0.40$ & $55.40 \pm 0.23$ \\
GraphEdit & $89.32 \pm 0.21$ & - & $85.38 \pm 0.07$ & - \\
SEGSL & $87.31 \pm 0.24$ & $63.53 \pm 0.08$ & $79.26 \pm 0.67$ & OOM \\
RCL & $86.20 \pm 1.00$ & $64.98 \pm 0.31$ & $78.66 \pm 0.26$ & $74.08 \pm 0.33$ \\
LLM4RGNN & - & - & $86.21 \pm 0.13$ & - \\
\midrule
\mname & $90.70 \pm 0.43$ & $67.23 \pm 0.09$ & $92.14 \pm 1.31$ & $76.87 \pm 0.62$\\
\bottomrule
\end{tabular}
\end{table*}

\subsubsection{\textbf{GSLM Learning Phase.}}  During the graph structure learning phase, our objectives are twofold: (1) to maximize the expected log joint likelihood of the observed and unobserved labels under the variational distribution $q(\bm{y}_U\mid\bm{s}_U)$ introduced in \Cref{eq:ELBO}, while keeping the parameters of the language model \( q_{\theta} \) fixed, and (2) to refine the input graph structure.

We have leveraged the LM to generate node representations \({h}_n \) for each node \( n \) in the graph. 
These embeddings capture rich semantic information, forming the input features for GSLM. 
In the task-specific prediction phase, inspired by \cite{besag1975statistical}, the pseudo-likelihood framework $\mathbb{E}_{q(y_U \mid s_U)} \left[ \log p(y_L, y_U \mid \{\bm{s}_n\}_{n \in \mathcal{V}}, A) \right]$, which we aim to optimize, can be approximated as:
\begin{equation}
    \max_\phi \mathbb{E}_{q(y_U \mid s_U)} \left[ \sum_{m \in \mathcal{V}} \log p(y_m \mid \{\bm{s}_n\}_{n \in \mathcal{V}}, A, y_{\mathcal{V} \setminus m}) \right],
\end{equation}
the expectation over the variational distribution \( q_\theta \) depends on the unobserved labels of the unlabeled nodes. To approximate this expectation, we generate pseudo-labels \( \hat{\bm{y}}_U \) by sampling from \( q_\theta(\bm{y}_U \mid \bm{s}_U) \). For each unlabeled node \( m \in U \), the pseudo-label \( \hat{\bm{y}}_m \) is drawn directly from the language model \( q_\theta \), facilitating efficient integration of semantic information into the graph learning process.

With the node representations \( \bm{h}_n \) and pseudo-labels \( \hat{\bm{y}}_U \), the pseudo-likelihood can be approximated, leading to the following objective function for optimizing the parameters \( \phi \):
\begin{align}
\mathcal{L}(\phi) &= \beta \sum_{m \in U} \log p_\phi\left( \hat{\bm{y}}_m \mid \{\bm{h}_n\}_{n \in \mathcal{V}}, \bm{A}, \bm{y}_L, \hat{\bm{y}}_{U \setminus m} \right) \nonumber \\
&\quad + (1 - \beta) \sum_{m \in L} \log p_\phi\left( \bm{y}_m \mid \{\bm{h}_n\}_{n \in \mathcal{V}}, \bm{A}, \bm{y}_{L \setminus m}, \hat{\bm{y}}_U \right),
\label{eq:gnn_objective}
\end{align}

where \( \beta \in [0,1] \) is a hyperparameter balancing the two terms.
\( \hat{\bm{y}}_{U \setminus m} \) denotes the pseudo-labels of all unlabeled nodes except node \( m \).
\( p_\phi(\bm{y}_m \mid \cdot) \) represents the GSLM's predictive distribution for node \( m \).

\noindent
\textbf{Graph structure refinement.}
With the initial adjacent matrix $\bm{A}$,
we propose two strategies to refine the graph structure, depending on the specific needs of the task.

\underline{\textit{Strategy 1: }}Since \mname operates within an iterative framework, LM and \gslm mutually improve each other's performance through maximizing the ELBO. 
Additionally, the graph structure used in the GSLM can be iteratively updated during the LM phase, leading to continuous refinement of both components. 
In the first category of graph structure refinement, the focus remains on optimizing $p_{\phi}$, where graph structure modifications are captured by implicit adjustments to the node features, ultimately influencing the final graph representations.
A similar idea is presented by Fang et al. \cite{fang2024universal}, who theoretically demonstrate that it is always possible to learn appropriate node features to perform any graph-level transformation, such as ``changing node features'' or ``adding or removing edges/subgraphs,'' through a learnable projection head.

\underline{\textit{Strategy 2: }}In the second category of graph refinement methods, we extend our approach by incorporating a graph structure optimization objective. Specifically, we propose a joint optimization method to learn both the graph structure and \gslm parameters by minimizing a hybrid loss function, which combines task-specific loss with graph refinement loss in the \gslm learning phase, denoted as $\mathcal{L}_{\text{GSLM}}=\mathcal{L}(\phi) + \mathcal{L_G}$.
Given the extensive success of existing graph structure learning works based on GNNs and the flexibility of \mname, we can seamlessly adapt the \gslm backbone as needed ($e.g.$, IDGL \cite{chen2020iterative}).
The specific graph refinement loss $\mathcal{L_G}$ utilized in \mname varies depending on the choice of the \gslm backbone architecture.
Incorporating the general loss function that jointly optimizes both GNN parameters and the underlying graph structures of the chosen \gslm backbone, the final loss function for \mname during the graph structure learning phase is formalized as follows: 
\begin{align}
\mathcal{L}_{\text{GSLM}} &= \mathcal{L}(\phi) + \mathcal{L_G} \nonumber \\
&= \beta \sum_{m \in U} \log p_\phi\left( \hat{\bm{y}}_m \mid \{\bm{h}_n\}_{n \in \mathcal{V}}, \bm{A}, \bm{y}_L, \hat{\bm{y}}_{U \setminus m} \right) \nonumber \\
&\quad + (1 - \beta) \sum_{m \in L} \log p_\phi\left( \bm{y}_m \mid \{\bm{h}_n\}_{n \in \mathcal{V}}, \bm{A}, \bm{y}_{L \setminus m}, \hat{\bm{y}}_U \right) \nonumber \\
&\quad + \mathcal{L_G}. \label{eq:gSLM_objective2}
\end{align}
As demonstrated by the experimental results, the second category of graph structure refinement methods shows slightly better performance due to the additional consideration of the structure refinement loss. However, the first category methods provide a good balance between computational complexity and performance, making them suitable for scenarios where resource constraints are a concern. 
Further details are provided in Section \ref{sec:experiment}.

\section{Experiment}
\label{sec:experiment}

\begin{figure*}[h]
    \centering
    \includegraphics[width=0.85\linewidth]{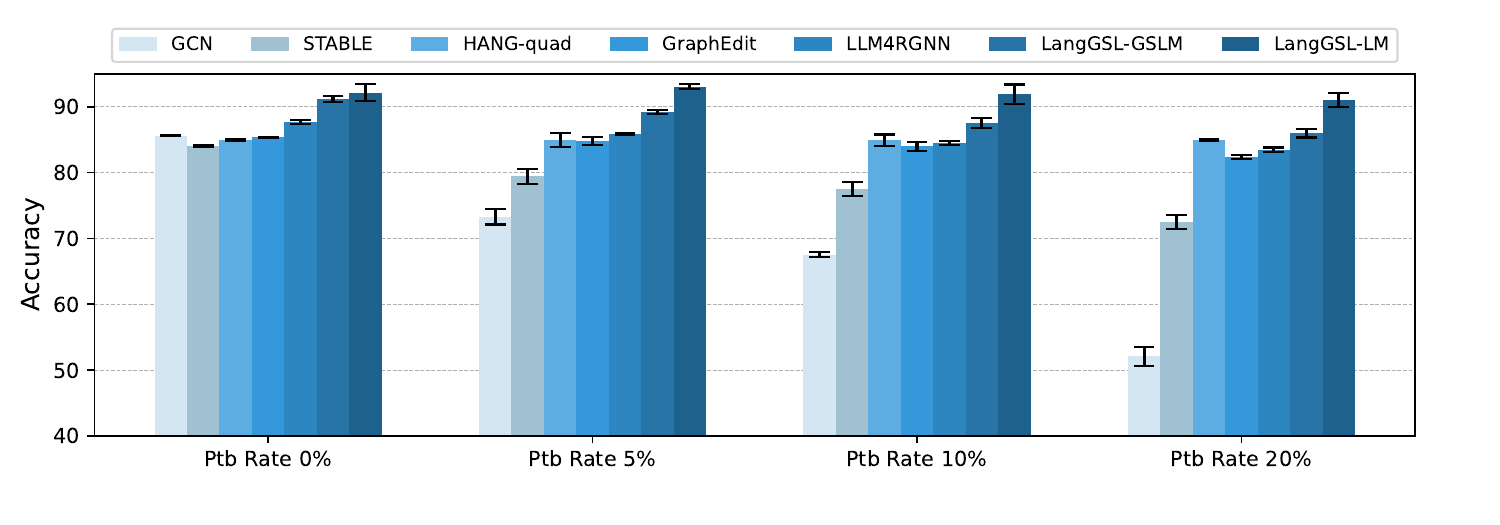}  
    \caption{Node classification accuracy (\%) on Pubmed under adversarial attack at various perturbation rates (Ptb Rate).}
    \label{fig:attPtb}
\end{figure*}


In this section, we conduct experiments to evaluate the proposed \mname framework across various task scenarios, aiming to address several key research questions:
\begin{enumerate}
    \item[\textbf{RQ1:}] How does \mname perform compared to existing GSLMs in the transductive learning setting, utilizing the default graph structure?
    \item[\textbf{RQ2:}] How robustness is \mname in the absence of graph structure or when facing graph topology attacks?
    \item[\textbf{RQ3:}] How does \mname compare to GraphLLM models in node classification tasks on TAGs?
    \item[\textbf{RQ4:}] How adaptable and flexible is \mname across different settings of backbones?

\end{enumerate}

\subsection{Experimental Setup}
\textbf{Datasets.} Four graph node classification benchmark datasets are included in our experiment, including ogbn-arxiv \cite{hu2020open}, Instagram \cite{huang2024can}, Pubmed, and Cora \cite{mccallum2000automating}.

\noindent
\textbf{Baselines. }
We evaluate the performance of our \mname model against a comprehensive set of the state-of-the-art baselines, spanning four key categories:
\begin{itemize}
\item \textbf{Graph Structure Learning Models (GSLMs): }
The baseline GSLMs considered include:
LDS \cite{franceschi2019learning}, GRCN \cite{yu2021graph}, ProGNN \cite{jin2020graph}, IDGL \cite{chen2020iterative}, GEN \cite{wang2021graph}, CoGSL \cite{liu2022compact}, SUBLIME \cite{liu2022towards}, 
SEGSL \cite{zou2023se},
STABLE \cite{li2022reliable}, NodeFormer \cite{wu2022nodeformer}, RCL \cite{zhang2024curriculum}, GraphEdit \cite{guo2024graphedit}, LLM4RGNN \cite{zhang2024can}, SLAPS \cite{fatemi2021slaps}, HES-GSL \cite{wu2023homophily},
and HANG-quad \cite{zhao2024adversarial}.

\item \textbf{Large Language Models for Graphs (GraphLLM):} GraphLLM focuses on leveraging the vast knowledge of LLMs to enhance graph learning: \textit{LLM-as-enhancer} (TAPE \cite{he2023harnessing}, GIANT \cite{chien2021node}, OFA \cite{liu2023one}), \textit{LLM-as-predictor} (GraphAdapter \cite{huang2024can}, LLaGA \cite{chen2024llaga}, GraphText \cite{zhao2023graphtext}, InstructGLM \cite{ye2023natural}), and \textit{LLM-as-aligner} (GLEM \cite{zhao2022learning}, PATTON \cite{jin2023patton}).

    \item \textbf{Pre-trained Language Models (PLMs):} The evaluation based on a diverse set of models, including BERT \cite{devlin2018bert}, DeBERTa \cite{he2020deberta}, RoBERTa \cite{liu2019roberta}, Sent-BERT \cite{reimers2019sentence}.

    \item \textbf{Graph Neural Networks (GNNs):} The baselines include GCN \cite{kipf2016semi}, GAT \cite{velivckovic2017graph}, GraphSAGE \cite{hamilton2017inductive}, and RevGAT \cite{li2021training}.

\end{itemize}
Notably, we consider the transductive node classification task in both the \textbf{T}opology \textbf{R}efinement (\textbf{TR}) scenario and the \textbf{T}opology \textbf{I}nference (\textbf{TI}) scenario, where the original graph structure is available for each method in \textbf{TR} but not in \textbf{TI}, with \textbf{TR} as the default setting, if there is no special instructions.
Further experimental and training details, including introductions and statistics of the TAG datasets, additional results, and implementation specifics can be found in Appendix \ref{app:dataset}, \ref{app:prompt}, 
 \ref{app:results}, and \ref{app:implementation} respectively.

\subsection{Overall Performance Comparison (RQ1)}

Table \ref{tab:gslWStr} presents the accuracy and standard deviation comparison under the transductive node classification task in the TR scenario, where the original graph structure is available for each method.
Based on the results, we have several observations:
\textit{\underline{Observation 1:} \textbf{
\mname demonstrates significant improvements over state-of-the-art GSLMs across datasets.}}
\mname consistently surpasses all baseline models, achieving an average improvement of $3.1\%$ compared to the second-best performance across all datasets. 
Notably, on Pubmed, \mname exhibits a substantial near $15\%$ improvement over the vanilla GCN, outperforming other methods by an even greater margin. 
These results validate \mname's effectiveness across both heterophilous and homophilous graphs, as well as its ability to handle graph structures of varying scales.
\textit{\underline{Observation 2:} \textbf{Leveraging abundant unlabeled data is essential for improving GSLM performance.}}
As mentioned in the recent literature \cite{fatemi2021slaps, li2024gslb}, optimizing graph structures solely
based on label information is insufficient. Leveraging a large and abundant amount of unlabeled
information can enhance the performance of GSLM ($e.g.,$ STABLE).
This confirms the effectiveness of the mutual learning mechanism in \mname, as validated by the results, which leverages the exchange of pseudo labels for unlabeled nodes to optimize LM and GSLM objectives and enhance learning performance.
\textit{\underline{Observation 3:} \textbf{Leveraging language models enhance the overall performance.}}
It is evident that methods leveraging language models (GraphEdit, LLM4RGNN, and \mname) significantly outperform traditional approaches primarily based on vanilla GNNs, leading to notable performance improvements.
What is more, as demonstrated in Appendix Figure \ref{fig:rawclean_study}, LLM in \mname enhances performance by effectively cleaning raw texts, leading to improved outcomes for downstream models.

\subsection{Robustness Analysis (RQ2)}

To investigate the robustness of the \mname\ algorithm, we primarily focus on the \textbf{TI} scenario where no graph structure is provided, and adversarial attack settings involving perturbed graph topology.

Table \ref{tab:gslWOStr} presents the results of the transductive node classification task under the \textbf{TI} scenario. 
Since certain GSLMs, such as IDGL \cite{chen2020iterative}, are originally designed for the \textbf{TR} scenario, we use \text{kNN} graphs as a substitute for their original graph structure.
It is evident that our proposed method, \mname, significantly outperforms other algorithms through all datasets.
Notably, LangGSL (LM) achieves an improvement of $16.37\%$ on Pubmed and $17.16\%$ on ogbn-arxiv over the second-best method, validating the ability of the LM component to predict reliable labels and provide a solid initial graph structure for the GSLM. In turn, LangGSL (GSLM) demonstrates that the graph refinement process further boosts performance, with improvements of $16.21\%$ on Pubmed and $4.02\%$ on ogbn-arxiv compared to the next best result.
Specifically, the Pubmed result even surpasses the performance in the \textbf{TR} setting, highlighting the robustness of LangGSL in scenarios where the original graph structure is not available.
In contrast, traditional methods such as GCN\textsubscript{kNN} and GSLMs like IDGL\textsubscript{kNN} display relatively lower accuracy, emphasizing the benefits of the mutual enhancement mechanism.

In the adversarial attack setting, we consider the state-of-the-art non-targeted attack method, Mettack \cite{DBLP:journals/corr/abs-1902-08412}, where the attacker has full access to node attributes, graph structure, and training labels. 
This allows the attacker to disrupt the entire graph topology and degrade the performance of graph structure learning models on the test set.
Referring to Figure \ref{fig:attPtb}, as the perturbation rate increases, the performance of all methods generally declines, indicating the adverse effect of noise on graph structure. 
However, both LangGSL-GSLM and LangGSL-LM demonstrate a remarkable level of robustness across all perturbation rates, outperforming existing robust GNN frameworks like LLM4RGNN, which is specifically designed to improve adversarial robustness in GNNs using LLMs. 
This further validates LangGSL's effectiveness in adversarial settings.

\subsection{Comparison with Large Language Models for Graphs (RQ3)}
We compare \mname with GraphLLM and methods that utilize only GNNs or LLMs on TAGs, as presented in Table \ref{tab:TAGs}.
To ensure a fair comparison and minimize the influence of different backbone models used in various GraphLLM methods, we followed the latest work \cite{li2024glbench} and implemented all methods using the same GNN (GCN \cite{kipf2016semi}) and LM (Sent-BERT \cite{reimers2019sentence}) backbones in our experiments.
LangGSL consistently ranks at or near the top across all three datasets compared with state-of-the-art methods with enhanced features provided by LLMs.
This reinforces the necessity of jointly considering both structure learning and node feature learning, as seamlessly integrated within \mname.
In the category of LLM-as-aligner, we observe that five out of the six top-performing results, including both the best and second-best, are achieved within this category.
This further reflects the effectiveness of the mutual learning between the two models in \mname.

\begin{table}[t]
\small
\centering
\caption{Accuracy $\pm$ STD comparison (\%) under the standard setting of the transductive node classification task in the Topology Inference (TI) scenario, which means the original graph structure is not available for each method.
"OOM" denotes out of memory. 
}
\label{tab:gslWOStr}
\begin{tabular}{lccc}
\toprule
 & \multicolumn{1}{c}{Pubmed} & \multicolumn{1}{c}{ogbn-arxiv} & \multicolumn{1}{c}{Instagram} \\
\midrule
MLP & $73.00 \pm 0.30$ & $55.21 \pm 0.11$ & $63.41 \pm 0.40$ \\
$\textnormal{GCN}_{knn}$ & $69.23 \pm 0.49$ & $55.21 \pm 0.22$ & $63.71 \pm 0.30$ \\
$\textnormal{GRCN}_{knn}$ & $68.96 \pm 0.85$ & OOM & $63.59 \pm 0.12$ \\
$\textnormal{GPRGNN}_{knn}$ & $68.19 \pm 0.19$ & OOM & $57.75 \pm 0.16$ \\
$\textnormal{IDGL}_{knn}$ & $74.01 \pm 0.64$ & OOM & $63.93 \pm 0.20$ \\
$\textnormal{SLAPS}_{knn}$ & $74.50 \pm 1.47$ & $55.19 \pm 0.21$ & $63.43 \pm 0.21$ \\
$\textnormal{GEN}_{knn}$ & $69.76 \pm 1.53$ & OOM & $63.56 \pm 0.11$ \\
$\textnormal{CoGSL}_{knn}$ & OOM & OOM & $64.03 \pm 0.10$ \\
$\textnormal{ProGNN}_{knn}$ & OOM & OOM & $57.47 \pm 0.23$ \\
SUBLIME & $75.08 \pm 0.55$ & $55.57 \pm 0.18$ & $57.52 \pm 0.40$ \\
NodeFormer & $59.83 \pm 6.50$ & $55.37 \pm 0.23$ & $64.34 \pm 0.39$ \\
HES-GSL & $77.08 \pm 0.78$ & $56.46 \pm 0.27$ & $64.58 \pm 0.45$ \\
\midrule
\mname (GSLM) & $93.21 \pm 0.12$ & $59.21 \pm 0.38$ & $66.27 \pm 0.18$ \\
\mname (LM) & $93.37 \pm 0.15$ & $73.62 \pm 0.26$ & $65.96 \pm 0.26$ \\
\bottomrule
\end{tabular}
\end{table}

\begin{table}[t]
\small
\centering
\caption{Accuracy results (\%) under supervised learning setting. 
The \colorbox{cyan!30}{best} and \colorbox{lightgray}{second-best} results are highlighted with distinct background colors.} 
\label{tab:TAGs}
\begin{tabular}{lcccc}
\toprule
  &{Instagram} & {Pubmed} & {ogbn-arxiv} 
 \\
\midrule
GCN  & 63.79 & 78.98 & 71.82 \\
GAT & 63.74  & 76.93 & 71.85 \\
GraphSAGE & 63.75  & 76.79 & 71.88 \\
\midrule
Sent-BERT (22M) & 63.07 & 65.93 & 72.82 \\
BERT (110M) & 63.75  & 63.69 & 72.29 \\
RoBERTa (355M) & 63.57  & 71.25 & 72.94 \\
\midrule
GIANT & 66.01  & 76.89 & 72.04 \\
TAPE & 65.85  & 79.87 & 72.99 \\
OFA & 60.85 & 75.61 & 73.23 \\
\midrule
InstructGLM &  57.94 & 71.26 & 39.09 \\
GraphText &  62.64 & 74.64 & 49.47 \\
GraphAdapter & \colorbox{cyan!30}{67.40} & 72.75 & 74.45 \\
LLaGA & 62.94 & 52.46 & 72.78 \\
\midrule
GLEMGNN & 66.10 & 81.72 & \colorbox{lightgray}{76.43} \\
GLEMLLM &  65.00 & 79.18 & 74.03 \\
PATTON &  64.27 & \colorbox{lightgray}{84.28} & 70.74 \\
\midrule
\mname &\colorbox{lightgray}{67.23} &\colorbox{cyan!30}{92.14} & \colorbox{cyan!30}{76.87} \\
\bottomrule
\end{tabular}
\end{table}

\subsection{Flexible Study (RQ4)}
\label{expsec:flexStudy}
\begin{figure}[h!]
    \centering
    \includegraphics[width=\linewidth]{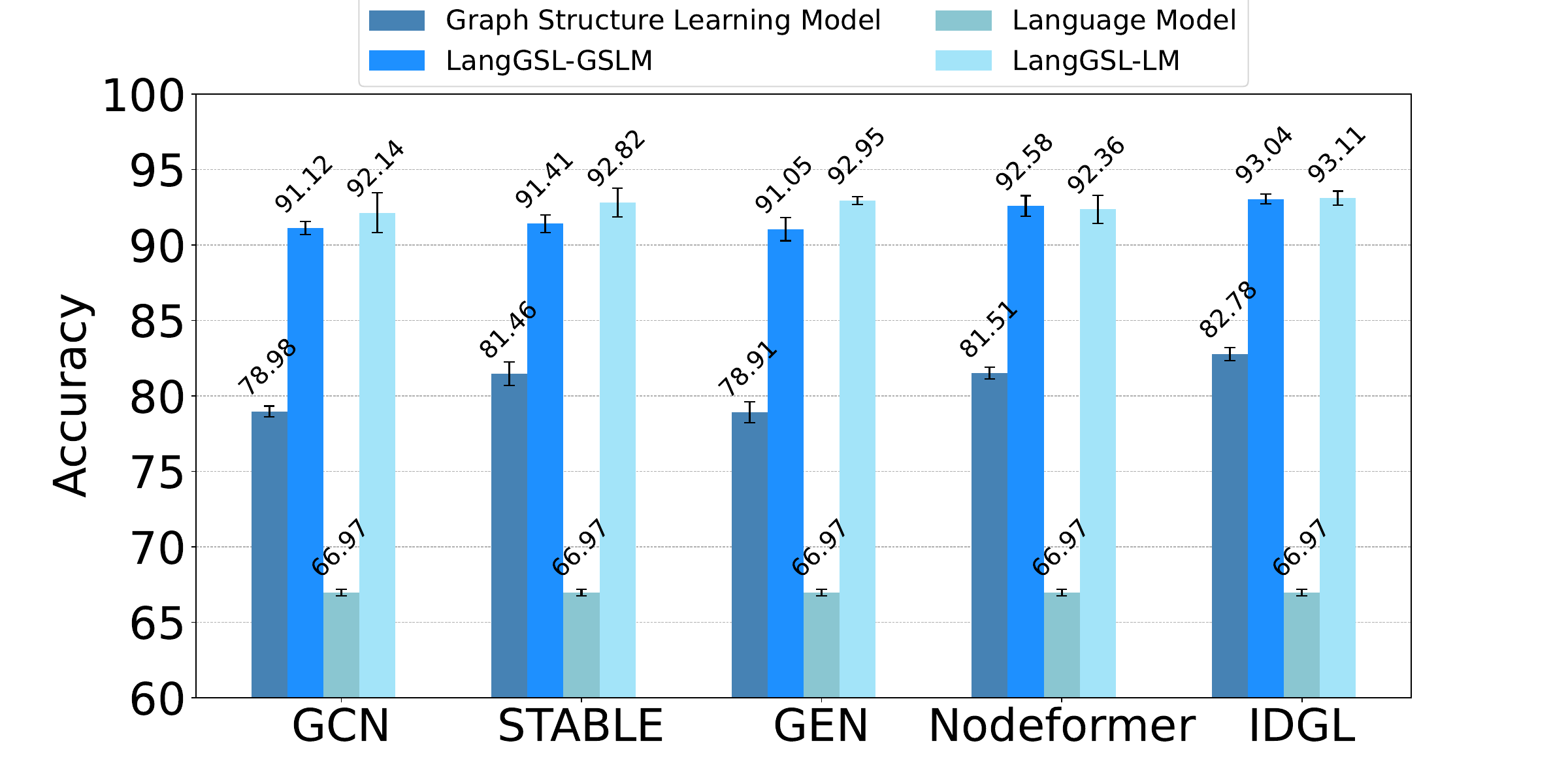}
    \caption{Experiments on Pubmed with different kinds of Graph Structure Learning backbone under \mname framework.}
    \label{fig:gslm_backbone}
\end{figure}

The GSLM component in \mname is designed to be flexible, allowing the choice of backbone to be adapted as needed. 
In this section, we compare the performance of \mname using various backbones, grouped by training procedure \cite{zhiyao2024opengsl}:
co-training (IDGL and NodeFormer), pre-training (STABLE), and iter-training (GEN), with GCN as the default baseline. 
Notably, as the backbone changes, the graph structure refinement loss is also adapted to align with the selected backbone, ensuring that the model remains optimized for the specific architecture.
As illustrated in Figure \ref{fig:gslm_backbone}, when integrated with LangGSL's mutual learning mechanism, GEN, STABLE, NodeFormer, and IDGL all further improve significantly upon their standalone performance.
Additionally, we observe a general trend where the more complex GSLMs, such as IDGL, achieve higher accuracy compared to the simpler GCN model.
Furthermore, the overall performance remains positively correlated with the inherent strength of the GSLMs.
Considering time and space constraints, reducing relatively small performance improvements for greater efficiency with GCN is a practical strategy, especially in resource-limited settings, where a trade-off between accuracy and computational cost is necessary.

Table \ref{tab:backbone} presents the experimental results on the Instagram dataset, showcasing performance across different GNN backbones and LLM scales.
Overall, the differences in the effects of combining different GNN and LM backbones are not substantial. 
In terms of LLM scaling, consistent with recent studies \cite{chen2024exploring, li2024glbench}, there are no definitive scaling laws for current approaches integrating LLMs with GNNs. 
Sent-BERT stands out as an efficient and effective baseline, offering an optimal balance between performance and computational cost.


\begin{table}[h!]
\scriptsize
    \centering
    \caption{Experiments on Instagram with different scales of LM and kinds of GNN backbone. }
    \label{tab:backbone}
    \begin{tabular}{lccc}
        \toprule 
        \multicolumn{2}{c}{\textbf{Backbone}} & \multicolumn{2}{c}{\textbf{Instagram}} \\
        & & \textbf{Acc} & \textbf{F1} \\
        \midrule
        Sent-BERT (22M) & GCN & ${67.23 \pm 0.09}$ & $57.03 \pm 0.72$  \\
        & RevGAT & $66.52 \pm 0.41$ & $54.89 \pm 1.73$ \\
        & GraphSAGE  & $66.59 \pm 0.16$ &$55.97 \pm 1.25$\\
        \midrule
        GCN & Sent-BERT (22M)  & $67.23 \pm 0.09$ & $57.03 \pm 0.72$  \\
        &DeBERTa (139M) & $66.64 \pm 0.31$ & $55.98 \pm 1.25$ \\
        &RoBERTa (335M) & $66.87 \pm 0.35$ & ${59.26 \pm 0.46}$ \\
        \bottomrule
    \end{tabular}
\end{table}

\section{Conclusion}

In this paper, we present \mname, a novel framework that integrates language models and GSLMs to address the key challenges in graph representation learning.
By leveraging LLMs for processing noisy node attributes, and harnessing the complementary strengths of LMs for generating task-related informative node embeddings, reliable graph structures, and pseudo labels, alongside GSLMs for refining graph structures and providing updated pseudo labels, LangGSL effectively considers both local and global graph information, achieving substantial improvements in node feature learning and structural optimization.
Through comprehensive experimental evaluations across diverse datasets, LangGSL consistently outperforms state-of-the-art methods, demonstrating its robustness in handling raw, noisy, incomplete, absent, and adversarial graphs.

\bibliographystyle{ACM-Reference-Format}

\newpage
\appendix
\newpage

\section{Deriving the Evidence Lower Bound (ELBO)}
\label{app:ELBO}
\begin{table*}[h!]
\small
\caption{Statistics of all datasets, including the number of nodes and edges, the average node degree, the average number of tokens in the textual descriptions associated with nodes, the number of classes, the training ratio in the supervised setting, the node text, the domains and tasks they belong to.}
\label{tabdatasets}
    \centering
    \begin{tabular}{lcccccccccc}
        \toprule
        \textbf{Dataset} & \textbf{\# Nodes} & \textbf{\# Edges} & \textbf{Avg. \# Deg} & \textbf{Avg. \# Tok} & \textbf{\# Classes} & \textbf{\# Train} & \textbf{Node Text} & \textbf{Domain}& \textbf{Group} \\
        \midrule
        
        Instagram & 11,339 & 144,010 & 25.40 & 59.25 & 2 & 10.00\% & User’s profile & Social &heterophilious\\
        Pubmed & 19,717 & 44,338 & 4.50 & 468.56 & 3 & 0.30\% & Paper content & Citation &homophilous\\
        ogbn-arxiv & 169,343 & 1,166,243 & 13.77 & 243.19 & 40 & 53.70\% & Paper content & Citation &homophilous\\
        Cora & 2,708 & 5,429 & 4.01 & 186.53 & 7 & 60.00\% & Paper content & Citation &homophilous\\
        \bottomrule
    \end{tabular}
\end{table*}

The primary objective in this work is to compute or maximize the log-likelihood of the observed labels given the node texts and graph structure:
\begin{equation}
    \log p(\bm{y}_L \mid \{\bm{s}_n\}_{n \in \mathcal{V}}, \bm{A}),
\end{equation}
where \(\bm{y}_L\) represents the observed labels, \(\bm{s}_V\) denotes the node texts, and \(\bm{A}\) is the graph structure. Here, directly computing this quantity is challenging due to the dependence on the unobserved labels \(\bm{y}_U\).

To address this challenge, we introduce a variational distribution \(q(\bm{y}_U \mid \bm{s}_U)\) to approximate the true posterior distribution \(p(\bm{y}_U \mid \{\bm{s}_n\}_{n \in \mathcal{V}}, \bm{A}, \bm{y}_L)\), thereby making the computation tractable.

Starting from the marginal likelihood, we seek to express the log-likelihood of the observed labels \(\bm{y}_L\) as:
\begin{equation}
\small
    \log p(\bm{y}_L \mid \{\bm{s}_n\}_{n \in \mathcal{V}}, \bm{A}) = \log \int p(\bm{y}_L, \bm{y}_U \mid \{\bm{s}_n\}_{n \in \mathcal{V}}, \bm{A}) \, \mathrm{d}\bm{y}_U.
\end{equation}
Since direct evaluation of this integral is intractable, we introduce the variational distribution \(q(\bm{y}_U \mid \bm{s}_U)\) to approximate the true posterior. By multiplying and dividing the integrand by \(q(\bm{y}_U \mid \bm{s}_U)\), we obtain:
\begin{equation}
\scriptsize
    \log p(\bm{y}_L \mid \{\bm{s}_n\}_{n \in \mathcal{V}}, \bm{A}) = \log \int q(\bm{y}_U \mid \bm{s}_U) \frac{p(\bm{y}_L, \bm{y}_U \mid \{\bm{s}_n\}_{n \in \mathcal{V}}, \bm{A})}{q(\bm{y}_U \mid \bm{s}_U)} \, \mathrm{d}\bm{y}_U.
\end{equation}
Applying Jensen's inequality, which exploits the concavity of the logarithmic function, gives:
\begin{equation}
\scriptsize
    \log p(\bm{y}_L \mid \{\bm{s}_n\}_{n \in \mathcal{V}}, \bm{A}) \geq 
    \int q(\bm{y}_U \mid \bm{s}_U) \log \left( \frac{p(\bm{y}_L, \bm{y}_U \mid \{\bm{s}_n\}_{n \in \mathcal{V}}, \bm{A})}{q(\bm{y}_U \mid \bm{s}_U)} \right) \, \mathrm{d}\bm{y}_U,
\end{equation}
which simplifies to:
\begin{equation}
\begin{split}
    \log p(\bm{y}_L \mid \{\bm{s}_n\}_{n \in \mathcal{V}}, \bm{A}) 
    &\geq \mathbb{E}_{q(\bm{y}_U \mid \bm{s}_U)} \left[ \log p(\bm{y}_L, \bm{y}_U \mid \{\bm{s}_n\}_{n \in \mathcal{V}}, \bm{A}) \right] \\
    &\quad - \mathbb{E}_{q(\bm{y}_U \mid \bm{s}_U)} \left[ \log q(\bm{y}_U \mid \bm{s}_U) \right].
\end{split}
\end{equation}

The right-hand side of this inequality is the Evidence Lower Bound (ELBO), which provides a tractable lower bound on the log-likelihood:
\begin{equation}
\begin{split}
    \text{ELBO} &= \mathbb{E}_{q(\bm{y}_U \mid \bm{s}_U)} \left[ \log p(\bm{y}_L, \bm{y}_U \mid \{\bm{s}_n\}_{n \in \mathcal{V}}, \bm{A}) \right] \\
    &\quad - \mathbb{E}_{q(\bm{y}_U \mid \bm{s}_U)} \left[ \log q(\bm{y}_U \mid \bm{s}_U) \right].
\end{split}
\end{equation}

Maximizing the ELBO with respect to both the variational distribution \(q(\bm{y}_U \mid \bm{s}_U)\) and the model parameters allows us to indirectly maximize the marginal log-likelihood \(\log p(\bm{y}_L \mid \{\bm{s}_n\}_{n \in \mathcal{V}}, \bm{A})\).

\section{Prompt design}
\label{app:prompt}

\begin{figure*}[h!]
    \centering
    \includegraphics[width=0.6\linewidth]{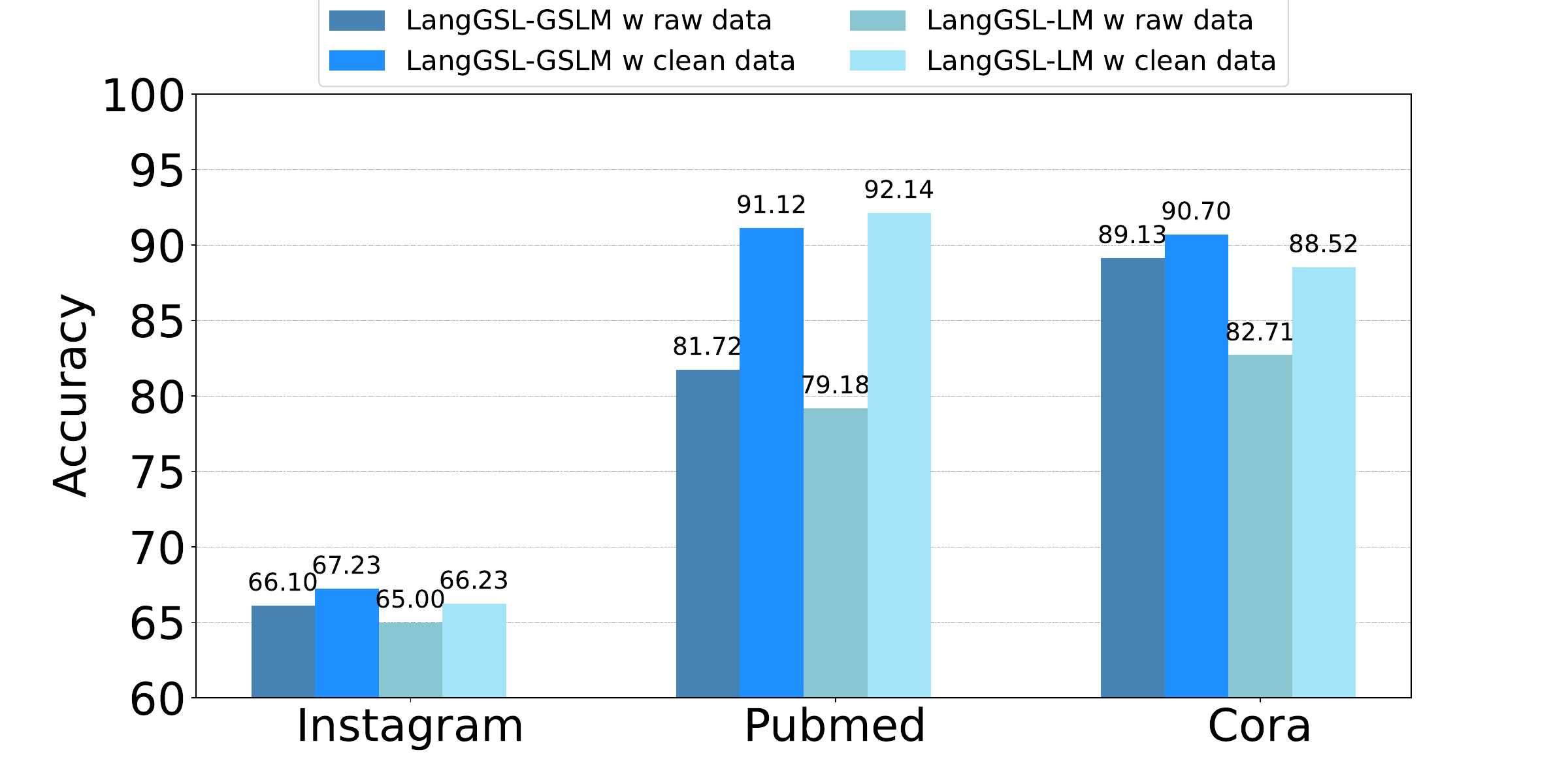}
    \caption{Comparison of test accuracy of \mname on Raw Texts vs. Cleaned Texts.}
    \label{fig:rawclean_study}
\end{figure*}

\textbf{Data cleaning verification.} {Figure \ref{fig:rawclean_study}} compares the test accuracy of \mname using raw texts versus cleaned texts across three datasets: Instagram, Pubmed, and Cora. The results clearly indicate that \mname consistently performs better when using cleaned texts, demonstrating the importance of data cleaning in text-attributed graphs (TAGs). For instance, on the Pubmed dataset, LangGSL-GSLM achieves an accuracy improvement of approximately 10.4\% when transitioning from raw to cleaned texts (81.72\% to 92.14\%). Similarly, LangGSL-LM shows a significant gain on Pubmed, increasing accuracy from 79.18\% to 91.49\%. Improvements on Instagram and Cora are also observed, though less pronounced. These findings validate the effectiveness of cleaning noisy raw texts in enhancing the overall model performance in downstream tasks.

\noindent
\textbf{Prompt sample.}
Building on the idea introduced in the Method section, we present a general example of a node $n \in \mathcal{V}$ from the Instagram dataset \cite{huang2024can}, including the prompt and the corresponding LLM (GPT-3.5-turbo \cite{ouyang2022training}) response: 
\begin{figure}[H]
    \centering
    \includegraphics[width=\linewidth]{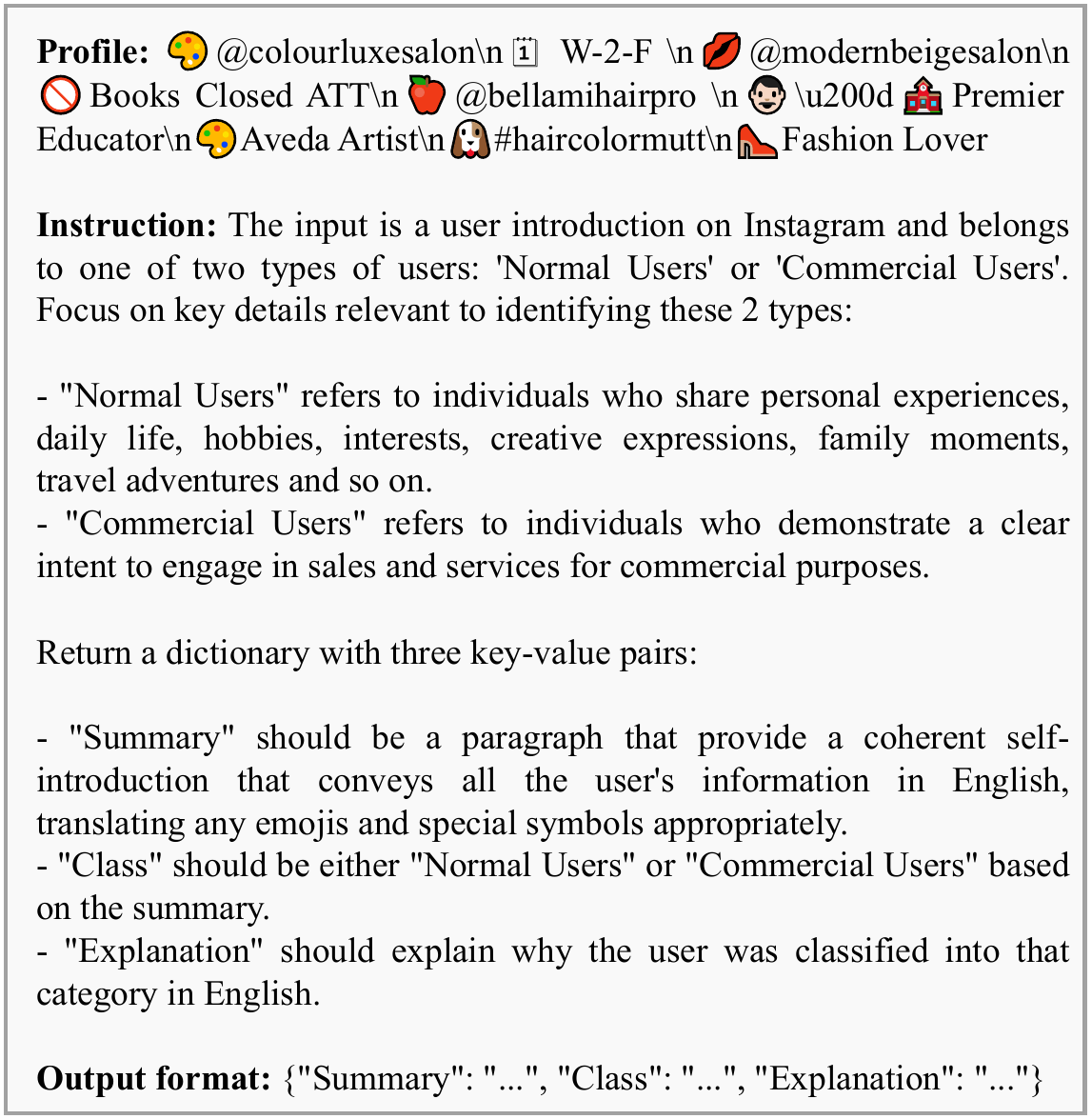}
    \vspace{-1em}  
    \includegraphics[width=\linewidth]{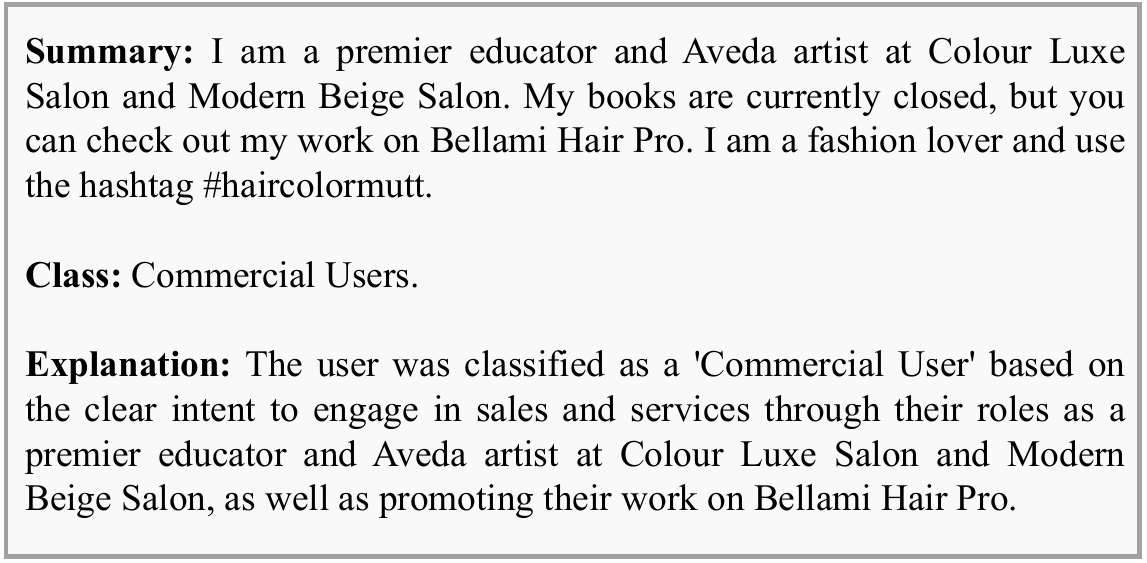}
    \label{fig:prompt_combined}
\end{figure}

\begin{figure*}[h!]
    \centering
    \includegraphics[width=\linewidth]{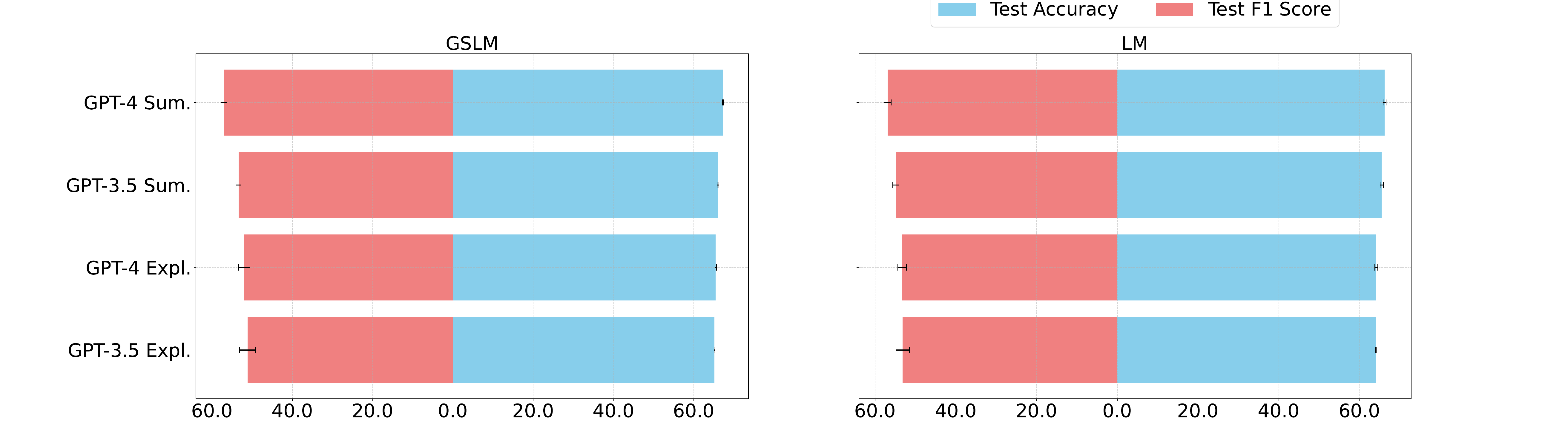}
    \caption{Test performance of \mname on the Instagram dataset using different prompts and LLMs settings. Results include test Accuracy (Acc) and F1 Score (F1). Each value is presented as mean ± standard deviation. Sum. means Summary, and Expl. means Explanation.}
    \label{fig:prompt_study}
\end{figure*}

\noindent
\textbf{Prompt type and LLM choice. }Figure \ref{fig:prompt_study} showcases the test performance of \mname on the Instagram dataset using different prompt types and LLM settings, comparing both test Accuracy (Acc) and F1 Score (F1) for GSLM and LM components. The models utilize summary-based (Sum.) and explanation-based (Expl.) outputs from GPT-4 and GPT-3.5. Notably, GPT-4 with summary-based prompts consistently achieves the highest accuracy and F1 scores across both GSLM and LM settings. This highlights the effectiveness of summary-based prompts in generating informative node representations and underscores the impact of prompt and model selection in enhancing performance.

\begin{figure}[H]
    \centering
    \includegraphics[width=\linewidth]{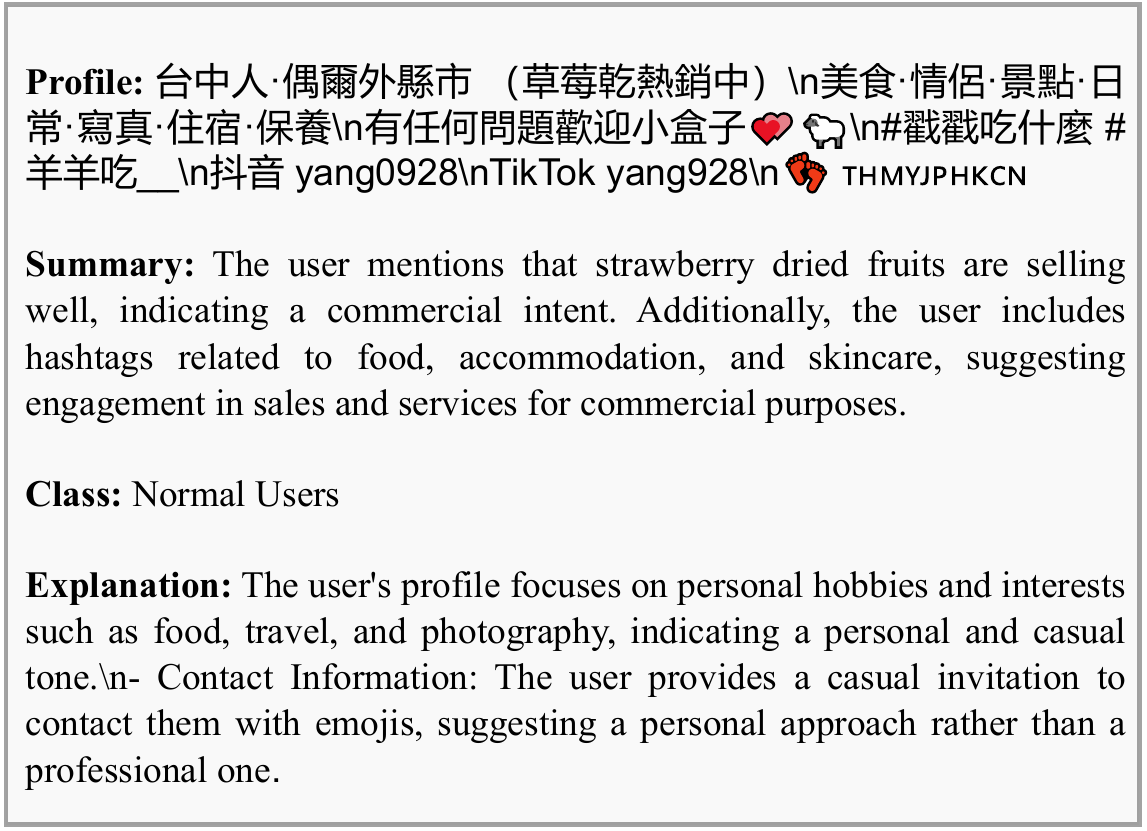}
    \label{fig:appromptanswer}
\end{figure}

\section{dataset}
\label{app:dataset}
We select four public and real-world datasets used for evaluation.
Table \ref{tabdatasets} shows detailed statistics of these datasets. 
And the details of text preprocessing and feature extraction
methods used for TAG datasets can be found in Table \ref{tabdatapreprocess}.
The datasets used in this work include Instagram, Pubmed, ogbn-arxiv, and Cora, each with distinct characteristics. Instagram is a social network dataset with heterophilous graph structures, consisting of user profile descriptions and two classes. Pubmed, ogbn-arxiv, and Cora are citation networks with homophilous graph structures, where nodes represent paper content.
Below are the
details of these datasets:

\noindent
\textbf{Pubmed. }The Pubmed dataset is a citation network comprising research papers and their citation relationships within the biomedical domain. The raw text data for the PubMed dataset was obtained from the GitHub repository provided by Chen et al. \cite{chen2024exploring}. In this dataset, each node represents a research paper, and each edge denotes a citation relationship between two papers.

\noindent
\textbf{ogbn-arxiv. }The ogbn-arxiv dataset is a citation network consisting of research papers and their citation relationships, collected from the arXiv platform. The raw text data for the Ogbn-arxiv dataset was obtained from the GitHub repository provided by OFA \cite{liu2023one}, with the original raw texts available \href{https://snap.stanford.edu/ogb/data/misc/ogbn_arxiv}{here}\footnote{\url{https://snap.stanford.edu/ogb/data/misc/ogbn_arxivs}}. In this dataset, each node represents a research paper, and each edge signifies a citation relationship.

\noindent
\textbf{Instagram.} The Instagram dataset is a social network where nodes represent users and edges represent following relationships. The raw text data for the Instagram dataset was collected from the GraphAdapter \cite{huang2024can}. The raw text associated with each node includes the user's personal introduction. Each node is labeled to indicate whether the user is classified as commercial or normal.

\noindent
\textbf{Cora.} The Cora dataset is a citation network comprising research papers and their citation relationships within the computer science domain. The raw text data for the Cora dataset was sourced from the GitHub repository provided by Chen et al. \cite{chen2024exploring}. In this dataset, each node represents a research paper, and the raw text feature associated with each node includes the title and abstract of the respective paper. An edge in the Cora dataset signifies a citation relationship between two papers. The label assigned to each node corresponds to the category of the paper.

\begin{table}[h!]
\small
\caption{Details of feature extraction methods used for TAG datasets. }
\label{tabdatapreprocess}
    \centering
    \begin{tabular}{lcccccccccc}
        \toprule
        \textbf{Dataset} & \textbf{Methods} & \textbf{\# Features} \\
        \midrule
        
        Instagram & TF-IDF & 500 \\
        Pubmed & TF-IDF & 500 \\
        ogbn-arxiv & skip-gram \cite{mikolov2013distributed} &  128 \\
        Cora & Bag-of-Words \cite{harris1954distributional} & 1433 \\
        \bottomrule
    \end{tabular}
\end{table}

\section{Additional results}
\label{app:results}
\textbf{Parameter study. }In the ELBO (Equation \ref{eq:ELBO}) optimization phase, we iteratively train LM and GSLM based on our derived equations: Equation \ref{eq:ELBO_LM} and Equation \ref{eq:gSLM_objective2}.
To adjust the relative influence of pseudo-labels in the learning process, we introduce two coefficients, $\alpha$ and $\beta$, denoted as LM-PL-Weight and GSLM-PL-Weight respectively. 
We then systematically analyze these two hyperparameters by examining how the performance varies with different values of $\alpha$ and $\beta$ on Pubmed dataset.

The parameter study of LangGSL demonstrates that both LangGSL-GSLM and LangGSL-LM maintain robust performance across different values of LM-PL-weight ($\alpha$) and GSLM-PL-weight ($\beta$). LangGSL-LM shows greater sensitivity to changes in these weights, with optimal performance achieved around the intermediate range of 0.3-0.5, while LangGSL-GSLM remains more stable. These findings highlight the importance of balanced pseudo-label contributions for maximizing the model's effectiveness.

\noindent
\textbf{Detailed results of \mname through the paper.}
Table \ref{tab:app-allresults} presents a comprehensive comparison of node classification accuracy (mean ± std\%) for \mname across all related downstream tasks, in comparison to baseline GNN- and LM-based methods. We include both shallow GNN baselines ($GNN_{\text{shallow}}$) and LangGSL's graph structure learning model (LangGSL-GSLM), alongside different LM-based methods, such as LLaMA2-7b (LLM), fine-tuned Sent-BERT (denoted as LM$_{\text{finetune}}$), and LangGSL-LM. The experiments are conducted across a variety of datasets, including Cora, Pubmed, Instagram, and ogbn-arxiv, under both standard and adversarial settings (with varying perturbation rates). The results consistently demonstrate that \mname outperforms the baselines in both GNN and LM categories, confirming its effectiveness and robustness across different graph structures, scales, and task scenarios. 

\begin{figure*}[h!]
    \centering
    \includegraphics[width=0.6\linewidth]{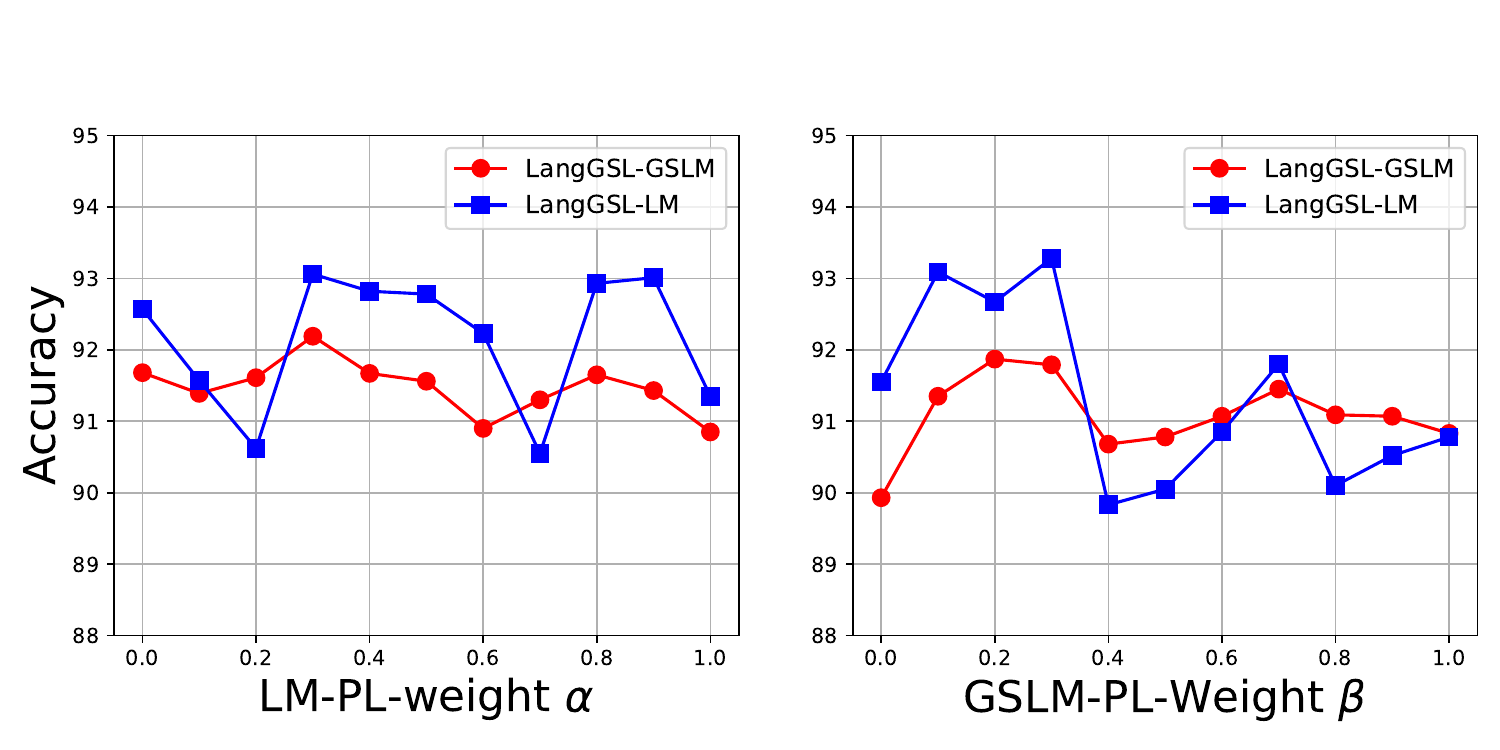}
    \caption{Parameter study of \mname.}
    \label{fig:exampl_figure}
\end{figure*}

\begin{table*}[h]

\small
    \centering
    \caption{Node classification accuracy (mean ± std\%) of \mname across all related downstream tasks discussed in this paper, compared with the node classification accuracy (mean ± std\%) of baseline GNN- and LM-based methods. 
    Fine-tuning Sent-BERT on labeled nodes, denoted as LM$_{\text{finetune}}$. Using embeddings generated by LLaMA2-7b denoted as LLM.  Using shallow features on GNN denoted as $GNN_{\text{shallow}}$.}
    \label{tab:app-allresults}
    \begin{tabular}{llccccccc}
        \toprule
        \multicolumn{1}{c}{Datasets} &\multicolumn{1}{c}{GSLM} & \multicolumn{2}{c}{$GNN$}  & \multicolumn{3}{c}{LM} \\
        \cmidrule(lr){3-4} \cmidrule(lr){5-7}
         && $GNN_{\text{shallow}}$
        & \textcolor{black}{\mname-GSLM}
        & LLM & LM$_{\text{finetune}}$ & \textcolor{black}{\mname-LM} \\
        \midrule
        &\textcolor{blue}{Table \ref{tab:gslWStr}}\\
        \midrule
        \multirow{1}{*}{Cora} 
        & GCN & $87.36\pm1.60$ & $90.70\pm0.43$ & $52.88\pm1.96$ & $76.07\pm3.78$ & $88.52\pm1.26$ \\
        \midrule
        \multirow{1}{*}{Pubmed} 
        & GCN & $78.98\pm0.35$ & $91.12\pm0.44$ & $66.07\pm0.56$ & $66.97\pm0.21$ & $92.14\pm1.31$ \\
        \midrule
        \multirow{1}{*}{Instagram} 
        & GCN & $63.79\pm0.08$ & $67.23\pm0.09$ & $62.47\pm0.98$ & $63.12\pm0.21$ & $66.23\pm0.38$ \\
        \midrule
        \multirow{1}{*}{ogbn-arxiv} 
        & GCN & $71.82\pm0.27$ & $76.87\pm0.62$ & $72.98\pm0.64$ & $72.80\pm0.13$ & $76.22\pm0.41$ \\
        \bottomrule
        &\textcolor{blue}{Table \ref{tab:gslWOStr}}\\
        \midrule
        \multirow{1}{*}{Pubmed} 
        & GCN & $ $ & $93.21\pm0.12$ &   &   & $93.37\pm0.15$ \\
        \midrule
        \multirow{1}{*}{ogbn-arxiv} 
        & GCN &   & $59.21\pm0.38$ &   &   & $73.62\pm0.26$ \\
        \midrule
        \multirow{1}{*}{Instagram} 
        & GCN &   & $66.27\pm0.18$ &  &   & $65.97\pm0.26$ \\
        \midrule
        &\textcolor{blue}{Figure \ref{fig:attPtb}}\\
        \midrule
        \multirow{1}{*}{Pubmed (ptb 0.00)} 
        & GCN &   & $91.12\pm0.44$ &   &  & $92.14\pm1.31$ \\
        \midrule
        \multirow{1}{*}{Pubmed (ptb 0.05)} 
        & GCN &   & $89.22\pm0.34$ &   &   & $93.03\pm0.37$ \\
        \midrule
        \multirow{1}{*}{Pubmed (ptb 0.10)} 
        & GCN &   & $87.56\pm0.79$ &   &   & $91.89\pm1.48$ \\
        \midrule
        \multirow{1}{*}{Pubmed (ptb 0.20)} 
        & GCN &   & $86.00\pm0.68$ &  &   & $91.04\pm1.05$ \\
        \midrule
        &\textcolor{blue}{Table \ref{tab:TAGs}}\\
        \midrule
        \multirow{1}{*}{Pubmed} 
        & GCN &   & $91.12\pm0.44$ &   &   & $92.14\pm1.31$ \\
        \midrule
        \multirow{1}{*}{Instagram} 
        & GCN &   & $67.23\pm0.09$ &   &   & $66.23\pm0.38$ \\
        \midrule
        \multirow{1}{*}{ogbn-arxiv} 
        & GCN &   & $76.87\pm0.62$&   &   & $76.22\pm0.41$ \\
        \midrule
        &\textcolor{blue}{Figure \ref{fig:gslm_backbone}}\\
        \midrule
        \multirow{4}{*}{Pubmed} 
        & GCN &   & $91.12\pm0.44$ &   &  & $92.14\pm1.31$ \\
        & STABLE &   & $91.41\pm0.60$ &   &   & $92.82\pm0.94$ \\
        & GEN &   & $91.05\pm0.77$ &   &   & $92.95\pm0.27$ \\
        & Nodeformer &   & $92.58\pm0.69$ &   &   & $92.36\pm0.93$ \\
        & IDGL &   & $93.04\pm0.32$ &   &   & $93.11\pm0.46$ \\
        \bottomrule
    \end{tabular}
\end{table*}

\section{implementation}
\label{app:implementation}

We perform comprehensive hyperparameter tuning to ensure a thorough and unbiased evaluation of both graph structure learning methods and LLM-based graph learning approaches.
To ensure the generality of the experimental results, all experiments were randomly run five times/seeds.
\textbf{The hyperparameter search
spaces of all methods are presented in Table \ref{aptab:paraGSL} and Table \ref{aptab:paraLLM}.}
We acknowledge and appreciate the comprehensive benchmarks provided by recent works \cite{li2024glbench, zhiyao2024opengsl, li2024gslb}, which have greatly informed and guided our work.

\noindent
\textbf{The configuration used in \mname: }
We conducted all experiments on an Nvidia A5000, using GPT-4 \cite{achiam2023gpt} for prompt engineering, GCN \cite{kipf2016semi} as the default GSLM backbone and Sent-BERT \cite{reimers2019sentence} as the default language model backbone unless otherwise specified.
The default GCN architecture consists of 2 hidden layers with 128 hidden units, a dropout rate of 0.5, a learning rate of 0.01, and a weight decay of 5e-4.



The  experiments across different datasets share following hyperparameter ranges:

\begin{itemize}
\item \textbf{GSLM} lr = {0.005, 0.01, 0.02}, layers = {2, 3}, hidden units = {64, 128, 256}, dropout = {0.2, 0.5, 0.6}, GSLM-pl-ratio = {0.2, 0.3, 0.5, 0.7, 0.8, 1.0}, GSLM-pl-weight = {0.2, 0.3, 0.5, 0.7, 0.8, 1.0}.

\item \textbf{LM} lr = {1e-5, 2e-5, 5e-5}, dropout = {0.1, 0.3}, cla-dropout = {0.1, 0.4}, LM-pl-ratio = {0.2, 0.3, 0.5, 0.7, 0.8, 1.0}, LM-pl-weight = {0.2, 0.3, 0.5, 0.7, 0.8, 1.0}.

\item \textbf{Iteration} inf-n-epochs = {1, 2, 3}.
\end{itemize}

These settings were used consistently across experiments to ensure the comparability of results.

\begin{table*}[t]
\centering
\scriptsize
\caption{Hyper-parameter search space of all implemented graph structure learning methods.}
\label{aptab:paraGSL}
\begin{tabular}{llp{7cm}}
    \toprule
    \textbf{Algorithm} & \textbf{Hyper-parameter} & \textbf{Search Space} \\
    \midrule
    \textbf{General Settings} & learning rate & 1e-2, 1e-3, 1e-4 \\
    & weight decay & 5e-3, 5e-4, 5e-5, 0 \\
    \midrule
    \textbf{GCN \cite{kipf2016semi}} & number of layers & 2, 3, 4 \\
    & hidden size & 16, 32, 64, 128 \\
    & dropout & 0, 0.2, 0.5, 0.8 \\
    \midrule
    \textbf{GRCN \cite{yu2021graph}} & \(K\) for nearest neighbors & 1, 5, 50, 100, 200 \\
    & learning rate for graph & 1e-1, 1e-2, 1e-3, 1e-4 \\
    \midrule
    \textbf{SLAPS \cite{fatemi2021slaps}} & learning rate of DAE & 1e-2, 1e-3 \\
    & dropout\_adj1 & 0.25, 0.5 \\
    & dropout\_adj2 & 0.25, 0.5 \\
    & \(k\) for nearest neighbors & 10, 15, 20 \\
    & \(\lambda\) & 0.1, 1, 10, 100, 500 \\
    & ratio & 1, 5, 10 \\
    & nr & 1, 5 \\
    \midrule
    \textbf{CoGSL \cite{liu2022compact}} & ve\_lr & 1e-1, 1e-2, 1e-3, 1e-4 \\
    & ve\_dropout & 0, 0.2, 0.5, 0.8 \\
    & \(\tau\) & 0, 0.2, 0.5, 0.8 \\
    & \(\epsilon\) & 0.1, 1 \\
    & \(\lambda\) & 0, 0.2, 0.5, 0.8, 1 \\
    \midrule
    \textbf{Nodeformer \cite{wu2022nodeformer}} & number of layers & 2, 3, 4 \\
    & hidden size & 32, 64, 128 \\
    & number of heads & 1, 2, 4 \\
    & dropout & 0, 0.2, 0.5, 0.8 \\
    & number of samples for gumbel softmax sampling & 5, 10, 20 \\
    & weight for edge reg loss & 1, 0.1, 0.01 \\
    \midrule
    \textbf{GEN \cite{wang2021graph}} & \(k\) for nearest neighbors & 8, 9, 10 \\
    & tolerance for EM & 1e-2, 1e-3 \\
    & threshold for adding edges & 0.5, 0.6, 0.7 \\
    \midrule
    \textbf{SEGSL \cite{zou2023se}} & \(K\) & 2, 3 \\
    & \(se\) & 2, 3 \\
    \midrule
    \textbf{LDS \cite{franceschi2019learning}} & \(\tau\) & 5, 10, 15 \\
    \midrule
    \textbf{ProGNN \cite{jin2020graph}} & learning rate for adj & 1e-1, 1e-2, 1e-3, 1e-4 \\
    \midrule
    \textbf{IDGL \cite{chen2020iterative}} & number of anchors & 300, 500, 700 \\
    & number of heads in structure learning & 2, 4, 6, 8 \\
    & \(\lambda_1\) for interpolation & 0.7, 0.8, 0.9 \\
    & \(\lambda_2\) for interpolation & 0.1, 0.2, 0.3 \\
    \midrule
    \textbf{SUBLIME \cite{liu2022towards}} & dropout rate for edge & 0, 0.25, 0.5 \\
    & \(\tau\) for bootstrapping & 0.99, 0.999, 0.9999 \\
    \midrule
    \textbf{STABLE \cite{li2022reliable}} & threshold for cosine similarity & 0.1, 0.2, 0.3 \\
    & \(k\) for nearest neighbors & 1, 3, 5 \\
    \midrule
    \textbf{RCL \cite{zhang2024curriculum}} & number of layers & 2, 3 \\
    & hidden size & 64, 128, 256 \\
    & \(\lambda\) & 1, 2, 3, 4, 5 \\
    & start \(\lambda\) & 5e-4, 1e-3, 2e-3 \\
    & node mask & 0, 1 \\
    & \(\beta\) & 0.95, 0.98, 1.00 \\
    \midrule
    \textbf{HES-GSL \cite{wu2023homophily}} & hidden size \(f\) & 64, 128, 256 \\
    & hidden size \(F\) & 256, 512, 1024 \\
    & pretrain epoch & 0, 200, 400 \\
    & loss weight & 1, 3, 5, 8, 10 \\
    & neighbor number & 3, 5, 10, 20, 30 \\
    & iteration step & 1, 3, 5, 8, 10 \\
    & noise weight & 5, 10, 20, 100 \\
    \bottomrule
\end{tabular}
\end{table*}

\begin{table*}[t]
\centering
\scriptsize
\caption{Hyper-parameter search space of all implemented LLM-based graph learning approaches.}
\label{aptab:paraLLM}
\begin{tabular}{llp{7cm}}
    \toprule
    \textbf{Algorithm} & \textbf{Hyper-parameter} & \textbf{Search Space} \\
    \midrule

    \textbf{DeBERTa \cite{he2020deberta}}
    & number of layers & 2, 3, 4 \\
    & hidden size & 64, 128 ,256 \\
    & dropout & 0.3, 0.5, 0.6 \\
    \midrule
    \textbf{RoBERTa \cite{liu2019roberta}}
    & number of layers & 2, 3, 4 \\
    & hidden size & 64, 128 ,256 \\
    & dropout & 0.3, 0.5, 0.6 \\
    \midrule
    \textbf{Sent-BERT \cite{reimers2019sentence}}
    & number of layers & 2, 3, 4 \\
    & hidden size & 64, 128 ,256 \\
    & dropout & 0.3, 0.5, 0.6 \\
    \midrule
    
    \textbf{TAPE \cite{he2023harnessing}}
    & number of layers & 2, 3, 4 \\
    & hidden size & 64, 128 ,256 \\
    & dropout & 0.3, 0.5, 0.6 \\
    \midrule
    \textbf{GIANT \cite{chien2021node}}
    & hidden size & 64, 128 ,256 \\
    & dropout & 0.3, 0.5, 0.6 \\
    \midrule
    \textbf{OFA Supervised \cite{liu2023one}}
    & learning rate & 1e-4 \\
    & \(JK\) & none \\
    & number of layers & 6 \\
    & hidden size & 768 \\
    & dropout & 0.15 \\
    \midrule
    \textbf{GraphAdapter \cite{huang2024can}}
    & learning rate & 5e-4 \\
    & number of layers & 5 \\
    & max length & 2048 \\
    & hidden size & 64, 128 \\
    \midrule
    \textbf{LLaGA \cite{chen2024llaga}}
    & learning rate & 2e-3 \\
    & warmup ratio & 0.03 \\
    & use hop & 2 \\
    & sample neighbour size & 10 \\
    & batch size & 16 \\
    & lr scheduler type & cosine \\
    & max length & 4096 \\
    \midrule
    \textbf{GraphText \cite{zhao2023graphtext}}
    & learning rate & 5e-5 \\
    & dropout & 0.5 \\
    & subgraph size & 3 \\
    & max length & 1024 \\
    \midrule
    \textbf{InstructGLM \cite{ye2023natural}}
    & learning rate & 8e-5 \\
    & warmup ratio & 0.05 \\
    & dropout & 0.1 \\
    & clip grad norm & 1 \\
    \midrule
    \textbf{GLEM \cite{zhao2022learning}}
    & number of layers & 2, 3, 4 \\
    & hidden size & 64, 128, 256 \\
    & dropout & 0.3, 0.5, 0.6 \\
    \midrule
    \textbf{PATTON \cite{jin2023patton}}
    & learning rate & 1e-5 \\
    & warmup ratio & 0.1 \\
    & dropout & 0.1 \\
    & batch size & 64 \\
    & max length & 256 \\
    \bottomrule
\end{tabular}
\end{table*}

\end{document}